\documentclass[10pt,twocolumn,letterpaper]{article}
\usepackage{cvpr}
\usepackage{multirow}
\usepackage{makecell}
\usepackage{array}
\usepackage{bm}
\usepackage{physics}
\usepackage{tikz}
\usetikzlibrary{positioning,arrows.meta,calc,fit,backgrounds,shapes.geometric,decorations.pathreplacing,patterns}
\usepackage{pgfplots}
\pgfplotsset{compat=1.17}
\usepackage{algorithm}
\usepackage{algpseudocode}
\usepackage{colortbl}
\usepackage{pifont}

\newcommand{\method}{Metric-VLA\xspace}
\newcommand{\simbench}{LIBERO-Pose\xspace}
\newcommand{\realbench}{ROP\xspace}
\newcommand{\realbenchfull}{Robot Object Pose in Real World\xspace}
\newcommand{\pizerofive}{$\pi_{0.5}$\xspace}

\newcommand{\prov}[1]{#1}

\newcommand{\cmark}{\ding{51}}

\newcommand{\frozenmark}{\ding{100}}
\definecolor{oursrow}{RGB}{232,240,252}
\newcommand{\ours}{\rowcolor{oursrow}}
\newcommand{\up}{$\uparrow$}
\newcommand{\down}{$\downarrow$}
\newcommand{\pz}{\phantom{0}}

\definecolor{cgoal}{RGB}{214,96,77}
\definecolor{cpose}{RGB}{67,147,195}
\definecolor{cact}{RGB}{90,174,97}
\definecolor{cgray}{RGB}{120,120,120}
\definecolor{cbg}{RGB}{246,246,246}

\newcommand{\R}{\mathbb{R}}

\newcommand{\SE}{\mathrm{SE}(3)}

\newcommand{\instr}{\ell}
\newcommand{\obs}{o}
\newcommand{\Iext}{I^{\mathrm{ext}}}
\newcommand{\Iwr}{I^{\mathrm{wr}}}
\newcommand{\Dext}{D^{\mathrm{ext}}}
\newcommand{\rstate}{\mathbf{s}}
\newcommand{\Tobj}{{\mathbf{T}}}
\newcommand{\Tgoal}{{\mathbf{G}}}

\newcommand{\Rm}{\mathbf{R}}

\providecommand{\pv}{}
\renewcommand{\pv}{\mathbf{p}}
\newcommand{\fB}{\mathcal{B}}
\newcommand{\fO}{\mathcal{O}}
\newcommand{\RBO}{\Rm_{\fB\fO}}

\newcommand{\rsix}{r_6}
\newcommand{\pscale}{\sigma_p}

\newcommand{\qtok}{\mathbf{q}}
\newcommand{\gtok}{\mathbf{g}}
\newcommand{\cq}{\mathbf{c}_q}
\newcommand{\cg}{\mathbf{c}_g}

\newcommand{\Achunk}{\mathbf{A}}

\newcommand{\temb}{\mathbf{e}(\tau)}

\newcommand{\modu}{\mathbf{u}}

\definecolor{linkblue}{rgb}{0.21,0.49,0.74}
\usepackage[pagebackref,breaklinks,colorlinks,allcolors=linkblue]{hyperref}
\usepackage[capitalize]{cleveref}

\providecommand{\ARXIVTW}{0.96}\providecommand{\ARXIVPRE}{-12pt}\providecommand{\ARXIVPOST}{2pt}\providecommand{\ARXIVBELOW}{-10pt}

\newcommand{\afflogo}[3]{\raisebox{#2}{\includegraphics[height=#1]{#3}}\hspace{0.25em}}
\title{Metric-VLA: Closing the Loop Between 3D Object Pose and Action}
\author{
Weihang Li$^{1,2}$\quad Junwen Huang$^{1,2,3,\dagger}$\quad Muayad Chalabi$^{1}$\quad Yuan Feng$^{1}$\quad Yuming Li$^{1,5}$\\
Junsheng Ding$^{1}$\quad Yuan Tian$^{4}$\quad Can Lin$^{1}$\quad Slobodan Ilic$^{1,5}$\quad Benjamin Busam$^{1,2}$\\
{\small $^{1}$\afflogo{0.80em}{-0.06em}{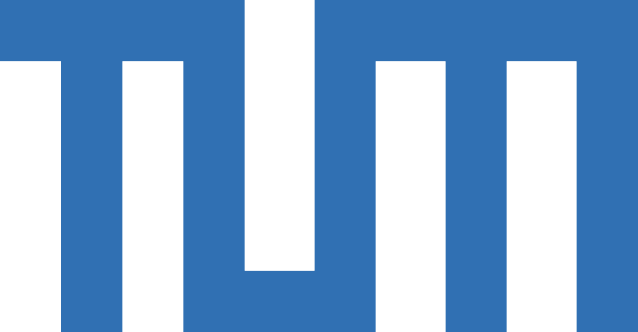}Technical University of Munich\quad $^{2}$\afflogo{0.50em}{0.08em}{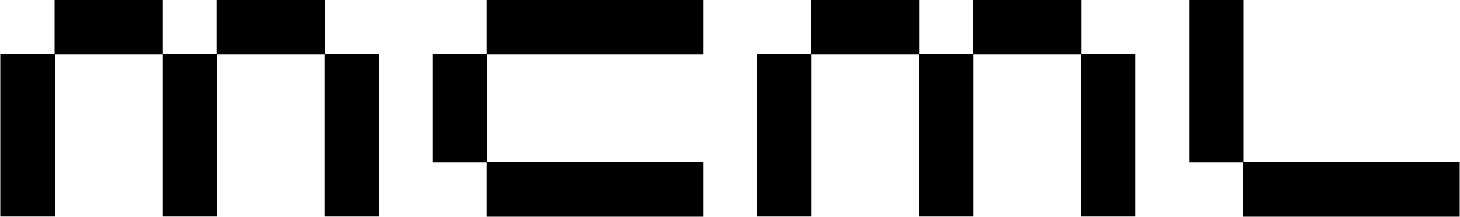}MCML\quad $^{3}$\afflogo{0.74em}{-0.08em}{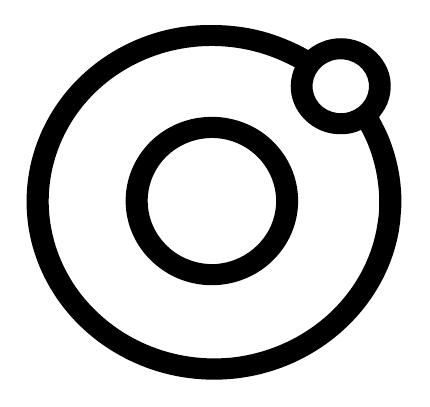}SavorCode\quad $^{4}$\afflogo{0.70em}{-0.02em}{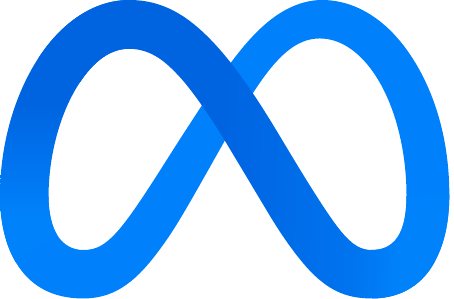}Meta\quad $^{5}$\afflogo{0.66em}{0.02em}{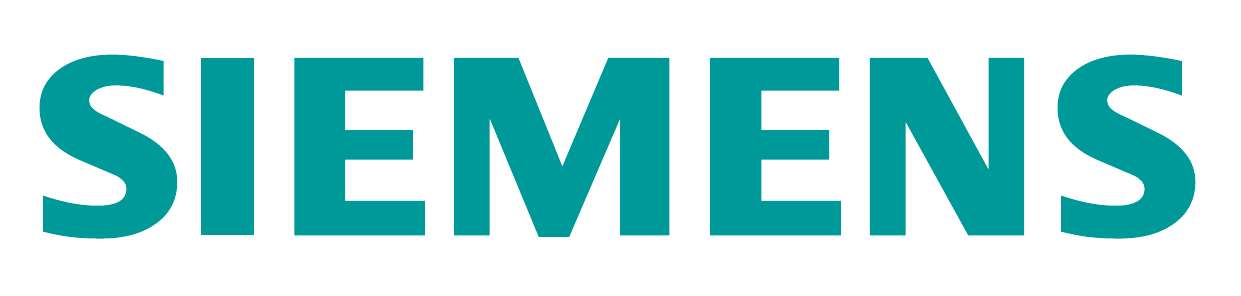}Siemens AG\qquad{\footnotesize $^{\dagger}$Corresponding author}}\\[-2pt]
{\footnotesize\texttt{\{weihang.li, junwen.huang, b.busam\}@tum.de}}
}
\newlength{\teaserwidth}

\begin{document}
\makeatletter
\g@addto@macro\@maketitle{%
  \vspace{\ARXIVPRE}%
  \begingroup
  \centering
  \captionsetup{hypcap=false}
  \ifdefined\teaserwidth\else\newlength{\teaserwidth}\setlength{\teaserwidth}{\textwidth}\fi%
\centering
\vspace{-16pt}
\pgfmathsetlengthmacro{\teaserunit}{\teaserwidth/988.243}
\begin{tikzpicture}[x=\teaserunit,y=-\teaserunit,inner sep=0pt]
  \node[anchor=north west] at (0,0) {\includegraphics[width=\teaserwidth,page=1]{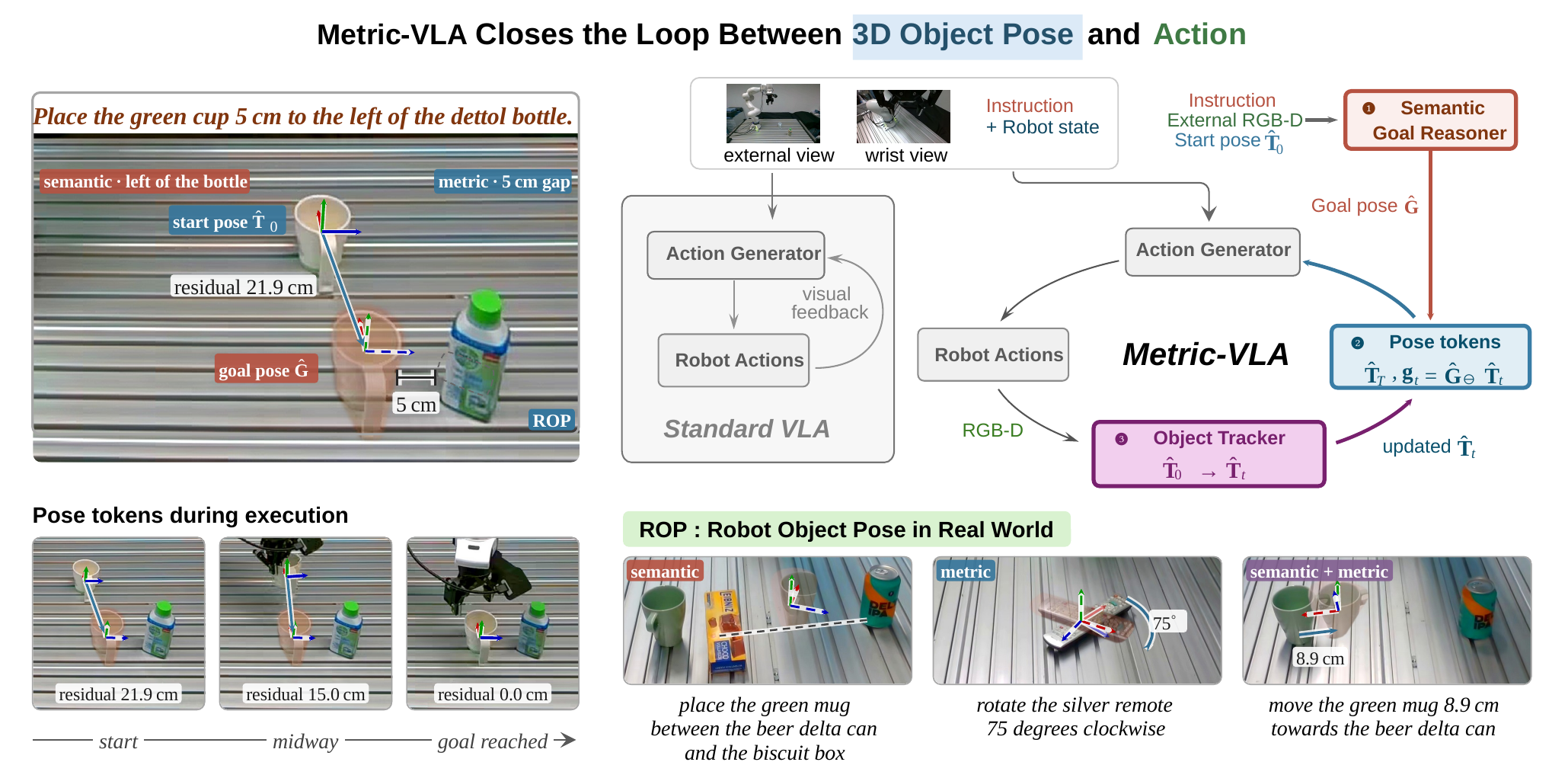}};
  \fill[fill={rgb,255:red,225;green,238;blue,245}] (858,226) rectangle (951,246);
  \pgfmathsetlengthmacro{\teaserstroke}{2.52279*\teaserunit}
  \draw[draw={rgb,255:red,57;green,125;blue,166},line width=\teaserstroke] (858,244.29) -- (951,244.29);
  \node[text={rgb,255:red,0;green,77;blue,111},font=\scriptsize] at (904,234.5)
    {\resizebox{0.088\teaserwidth}{!}{$\hat\Tobj_t,\;\gtok_t=\hat\Tgoal\ominus\hat\Tobj_t$}};
\end{tikzpicture}
\ifdefined\ARXIVBELOW\vspace{\ARXIVBELOW}\else
\vspace{-12pt}
\fi
\captionof{figure}{\textbf{\method bridges 3D object pose with VLA action generation, making the object manipulation metrically evaluable.}
\ding{182}~The semantic goal reasoner maps language to an object goal pose.
\ding{183}~Pose tokens condition the action expert on the current pose and goal residual.
\ding{184}~The object tracker updates the pose conditions after each action chunk.
\textbf{\realbench} is a new real-world robotic benchmark that scores semantic, metric, and composed manipulation using annotated object poses.}
\label{fig:teaser}

  \par
  \endgroup
  \vspace{\ARXIVPOST}
}
\makeatother
\maketitle
\begin{abstract}
Vision-language-action (VLA) policies learn diverse manipulation skills from large-scale, task-oriented datasets, yet often struggle with metric instructions such as ``move the bottle 10 cm to the right''. Following these commands requires tracking the object's motion and determining how much remains to reach the target, quantities that current VLAs typically leave implicit.
We present \method, which closes the loop directly between 3D object pose and action.
A goal reasoner combines visual grounding with geometric composition to translate the instruction into a 3D object goal pose. Two continuous pose tokens provide the action expert with the current pose of the commanded object and its remaining residual to the goal.
After each action chunk, a CAD-free tracker fuses visual tracks with gripper kinematics to update the object pose and goal residual. The consequence of acting becomes the next condition for planning.
We evaluate metric instructions, spatial relations, and their compositions using the achieved object pose. For this, we introduce LIBERO-Pose and ROP to benchmark simulated and real robot actions. On LIBERO-Pose, \method outperforms the strongest of seven learned baselines, raising overall success from 17.4\% to 71.3\%; on \realbench, it improves over \pizerofive{} from 24.0\% to 72.7\%.
Project page: \href{https://colin-de.github.io/Metric-VLA/}{colin-de.github.io/Metric-VLA}.
\end{abstract}
\section{Introduction}
\label{sec:intro}
Robotic manipulation has long served as the primary \emph{raison d'\^etre} for 3D object pose estimation~\cite{collet2011moped,xiang2018posecnn,tremblay2018dope,wang2019densefusion,jung2024housecat6d,zhang2024omni6dpose,wen2024foundationpose,Li_2026_CVPR}.
Despite this close connection, pose estimation and manipulation policy learning have largely progressed along separate paths.
Pose estimation is often evaluated on static, curated images~\cite{hodan2017t,hodan2018bop,brachmann2014learning}, with object pose treated as a perception output rather than a feedback state during physical execution.
In parallel, robot learning has advanced from visuomotor policies trained end to end~\cite{pomerleau1988alvinn,levine2016endtoend} and imitation policies~\cite{chi2023diffusion,zhao2023act}, to vision-language-action (VLA) models~\cite{zitkovich2023rt2,kim2024openvla,black2024pi0,intelligence2025pi05} and world action models~\cite{guo2024pad,yuan2026fastwam,kim2026cosmospolicy}, which generalize across diverse semantic tasks~\cite{intelligence2025pi05,chen2026steerable}.
Yet these models struggle with simple metric instructions such as \emph{``move the bottle 10\,cm to the right''} or \emph{``rotate the remote 75 degrees clockwise''}.
In our experiments, these policies frequently fail to reach the requested displacement or rotation, even after fine-tuning on task demonstrations.
These failures motivate a closer connection between estimating an object's pose and controlling how it changes.

Following a metric instruction requires grounding a target pose and tracking the motion still needed to reach it.
For relative commands, this means retaining the object's pose at command time as it moves.
Depth and point-cloud representations strengthen spatial grounding~\cite{qu2025spatialvla,sun2025geovla,chen2026pointact}, while systems such as Open6DOR~\cite{ding2024open6dor} and SoFar~\cite{qi2025sofar} translate language into target positions and orientations for grasping and motion planning.
Instead, \textbf{we bring this goal reasoning together with continuously updated object-state feedback within a single VLA framework.}

\begin{figure*}[t]
\centering
\begin{tikzpicture}[
  font=\scriptsize, x=1cm, y=1cm,
  hdr/.style={font=\footnotesize\bfseries, anchor=base west, inner sep=0pt},
  sym/.style={font=\scriptsize, inner sep=1pt},
  gsym/.style={font=\scriptsize, inner sep=1pt, text=black!55},
  blk/.style={draw=black!60, fill=white, line width=0.5pt, rounded corners=2pt, align=center, inner sep=4pt, font=\scriptsize},
  arr/.style={-{Stealth[length=1.6mm]}, line width=0.6pt, draw=black!65, shorten >=0.6pt},
  parr/.style={-{Stealth[length=1.6mm]}, line width=1.1pt, draw=cpose!75!black},
  garr/.style={-{Stealth[length=1.6mm]}, line width=1.1pt, draw=cgoal!85!black},
  tk/.style={draw=black!50, line width=0.3pt, minimum width=0.22cm, minimum height=0.30cm, inner sep=0pt},
  mlp/.style={fill=cpose!14, draw=cpose!80!black, line width=0.4pt},
  chip/.style={font=\scriptsize, inner sep=1.3pt, text=white, fill=black!65, fill opacity=0.9, text opacity=1, rounded corners=1pt},
  halo/.style={preaction={draw=white, line width=#1, opacity=0.8}}, halo/.default=2pt,
  tref/.style={draw=black!60, dash pattern=on 1.2pt off 0.9pt, line width=0.45pt},
  tseg/.style={dash pattern=on 2pt off 1.2pt, line width=0.6pt, draw=black!75, halo=1.6pt},
  tax/.style={-{Stealth[length=1.0mm,width=0.9mm]}, line width=0.7pt, halo=1.6pt},
  taxxs/.style={tax, draw=red!75!black}, taxys/.style={tax, draw=green!60!black}, taxzs/.style={tax, draw=blue!75!black},
  taxxg/.style={tax, draw=red!75!black, dash pattern=on 1.5pt off 1pt}, taxyg/.style={tax, draw=green!60!black, dash pattern=on 1.5pt off 1pt}, taxzg/.style={tax, draw=blue!75!black, dash pattern=on 1.5pt off 1pt},
  tlab/.style={font=\scriptsize, inner sep=1pt},
  taxs/.style={tax, draw=cpose!80!black}, tays/.style={tax, draw=cpose!80!black},
  taxg/.style={tax, draw=cgoal!85!black, dash pattern=on 1.5pt off 1pt}, tayg/.style={tax, draw=cgoal!85!black, dash pattern=on 1.5pt off 1pt},
]
\newcommand{\photocard}[5]{%
  \begin{scope}
    \clip[rounded corners=2pt] (#1,#2) rectangle ++(#3,#4);
    \node[inner sep=0pt, anchor=south west] at (#1,#2) {\includegraphics[width=#3cm,height=#4cm]{figures/teaser/#5.jpg}};
    \begin{scope}[shift={(#1,#2)}, x=#3cm, y=#4cm]\tikzset{
      tax/.style={-{Stealth[length=0.45mm,width=0.30mm]}, line width=0.42pt, line cap=round, line join=round, halo=0.70pt},
      taxxg/.style={tax, draw=red!75!black, dash pattern=on 0.8pt off 0.5pt},
      taxyg/.style={tax, draw=green!60!black, dash pattern=on 0.8pt off 0.5pt},
      taxzg/.style={tax, draw=blue!75!black, dash pattern=on 0.8pt off 0.5pt},
      tseg/.style={dash pattern=on 0.9pt off 0.7pt, line width=0.30pt, draw=black!60, halo=0.65pt},
      tref/.style={draw=black!40, dash pattern=on 0.8pt off 0.6pt, line width=0.25pt},
    }
    \input{figures/teaser/#5_overlay}\end{scope}
  \end{scope}
  \draw[draw=black!30, line width=0.5pt, rounded corners=2pt] (#1,#2) rectangle ++(#3,#4);}
\newcommand{\triad}[5]{\begin{scope}[shift={(#1,#2)}, rotate=#3]\draw[tax#4] (0,0) -- (#5,0); \draw[tay#4] (0,0) -- (0,#5); \fill[black!70] (0,0) circle (0.04);\end{scope}}
\newcommand{\mlpglyph}[2]{\path[mlp] (#1,#2-0.11) -- (#1+0.34,#2-0.17) -- (#1+0.34,#2+0.17) -- (#1,#2+0.11) -- cycle;}
\newcommand{\photolabel}[3]{%
  \foreach \a in {0,30,...,330} \node[tlab, #1, text=white, shift={(\a:0.5pt)}] at (#2) {#3};
  \node[tlab, #1] at (#2) {#3};}

\begin{scope}[on background layer]
  \fill[cgoal!5, rounded corners=3pt] (0,1.05) rectangle (5.40,5.50);
  \fill[cpose!6, rounded corners=3pt] (5.60,1.05) rectangle (9.70,5.50);
  \fill[cact!6, rounded corners=3pt] (9.90,1.05) rectangle (17.40,5.50);
\end{scope}
\node[hdr, text=cgoal!80!black] at (0.12,5.95) {(a) Semantic Goal Reasoning};
\node[hdr, text=cpose!80!black] at (5.72,5.95) {(b) Object Pose Tokens};
\node[hdr, text=cact!45!black] at (10.02,5.95) {(c) Pose-Conditioned Action Generation};
\node[gsym, anchor=base west, inner sep=0pt] at (0.12,5.63) {at command time};
\node[gsym, anchor=base west, inner sep=0pt] at (5.72,5.63) {updated after every chunk};
\node[gsym, anchor=base west, inner sep=0pt] at (10.02,5.63) {replanned after every chunk};

\node[font=\scriptsize, text width=5.0cm, align=center, anchor=north, inner sep=1pt] (instruction) at (2.70,5.38) {$\instr$: \emph{``place the green mug between the}\\[-1pt]\emph{beer delta can and the biscuit box''}};
  \begin{scope}
    \clip[rounded corners=2pt] (0.15,2.90) rectangle ++(1.62,1.12);
    \node[inner sep=0pt, anchor=south west] at (0.15,2.90) {\includegraphics[width=1.62cm,height=1.12cm]{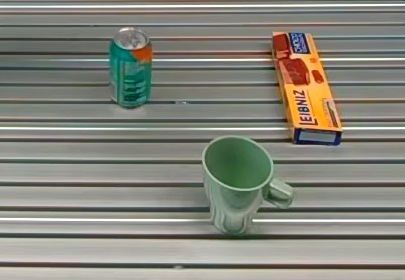}};
    \begin{scope}[shift={(0.15,2.90)}, x=1.62cm, y=1.12cm]\path[tref] (0.270,0.671) -- (0.269,0.667) -- (0.269,0.664) -- (0.269,0.663) -- (0.269,0.661) -- (0.269,0.660) -- (0.269,0.659) -- (0.269,0.658) -- (0.269,0.655) -- (0.269,0.654) -- (0.270,0.652) -- (0.271,0.647) -- (0.272,0.644) -- (0.272,0.644) -- (0.272,0.642) -- (0.273,0.641) -- (0.274,0.639) -- (0.275,0.637) -- (0.275,0.636) -- (0.277,0.634) -- (0.278,0.632) -- (0.280,0.629) -- (0.282,0.627) -- (0.283,0.625) -- (0.286,0.623) -- (0.288,0.620) -- (0.292,0.618) -- (0.294,0.616) -- (0.298,0.614) -- (0.302,0.612) -- (0.304,0.611) -- (0.306,0.610) -- (0.308,0.609) -- (0.315,0.608) -- (0.318,0.608) -- (0.320,0.608) -- (0.326,0.608) -- (0.327,0.608) -- (0.331,0.608) -- (0.332,0.608) -- (0.338,0.610) -- (0.340,0.611) -- (0.343,0.612) -- (0.345,0.613) -- (0.348,0.615) -- (0.352,0.618) -- (0.354,0.619) -- (0.357,0.622) -- (0.359,0.624) -- (0.361,0.626) -- (0.362,0.628) -- (0.363,0.630) -- (0.364,0.631) -- (0.365,0.634) -- (0.366,0.635) -- (0.366,0.636) -- (0.367,0.637) -- (0.368,0.640) -- (0.369,0.642) -- (0.370,0.646) -- (0.371,0.649) -- (0.372,0.654) -- (0.372,0.658) -- (0.372,0.662) -- (0.372,0.663) -- (0.372,0.664) -- (0.372,0.667) -- (0.371,0.668) -- (0.371,0.672) -- (0.369,0.678) -- (0.369,0.678) -- (0.369,0.679) -- (0.367,0.684) -- (0.365,0.687) -- (0.363,0.691) -- (0.361,0.693) -- (0.360,0.695) -- (0.359,0.696) -- (0.358,0.697) -- (0.357,0.698) -- (0.356,0.699) -- (0.354,0.701) -- (0.353,0.702) -- (0.349,0.704) -- (0.344,0.708) -- (0.341,0.709) -- (0.337,0.711) -- (0.332,0.713) -- (0.331,0.713) -- (0.328,0.714) -- (0.324,0.715) -- (0.322,0.715) -- (0.320,0.715) -- (0.318,0.715) -- (0.315,0.715) -- (0.315,0.715) -- (0.312,0.714) -- (0.308,0.714) -- (0.301,0.712) -- (0.300,0.711) -- (0.298,0.711) -- (0.294,0.708) -- (0.292,0.707) -- (0.288,0.704) -- (0.284,0.701) -- (0.281,0.697) -- (0.281,0.697) -- (0.280,0.696) -- (0.279,0.694) -- (0.277,0.692) -- (0.277,0.691) -- (0.277,0.691) -- (0.275,0.688) -- (0.274,0.686) -- (0.273,0.684) -- (0.273,0.683) -- (0.272,0.682) -- (0.272,0.679) -- (0.270,0.675) -- cycle;
\path[tref] (0.669,0.789) -- (0.716,0.509) -- (0.723,0.501) -- (0.724,0.499) -- (0.725,0.499) -- (0.726,0.499) -- (0.734,0.498) -- (0.799,0.495) -- (0.823,0.494) -- (0.826,0.494) -- (0.827,0.494) -- (0.830,0.495) -- (0.831,0.495) -- (0.835,0.497) -- (0.836,0.498) -- (0.835,0.502) -- (0.835,0.503) -- (0.831,0.523) -- (0.826,0.545) -- (0.819,0.578) -- (0.809,0.621) -- (0.789,0.707) -- (0.762,0.824) -- (0.758,0.839) -- (0.758,0.840) -- (0.758,0.840) -- (0.756,0.841) -- (0.747,0.844) -- (0.746,0.844) -- (0.743,0.844) -- (0.727,0.845) -- (0.695,0.847) -- (0.676,0.848) -- (0.667,0.849) -- (0.665,0.849) -- (0.661,0.846) -- (0.661,0.845) -- (0.661,0.843) -- (0.663,0.831) -- (0.664,0.824) -- (0.666,0.810) -- (0.666,0.808) -- cycle;
\draw[taxxs] (0.597,0.307) -- (0.421,0.441);
\draw[taxys] (0.597,0.307) -- (0.630,0.514);
\draw[taxzs] (0.597,0.307) -- (0.666,0.485);
\filldraw[draw=black!65, fill=white, line width=0.25pt] (0.597,0.307) circle[radius=0.45pt];\end{scope}
  \end{scope}
  \draw[draw=black!30, line width=0.4pt, rounded corners=2pt] (0.15,2.90) rectangle ++(1.62,1.12);
\node[sym, anchor=south] at (0.96,4.06) {$\Iext_0,\Dext_0$};
\photolabel{text=cpose!75!black, anchor=east}{0.99,3.18}{$\hat\Tobj_0$}
\node[blk, text width=1.15cm, inner xsep=2pt, fill=cgoal!10, draw=cgoal!80!black] (gr) at (2.70,3.46) {Goal\\Reasoner};
\draw[arr] (1.77,3.46) -- (gr.west);
\draw[arr] (instruction.south) -- (gr.north);
  \begin{scope}
    \clip[rounded corners=2pt] (3.63,2.90) rectangle ++(1.62,1.12);
    \node[inner sep=0pt, anchor=south west] at (3.63,2.90) {\includegraphics[width=1.62cm,height=1.12cm]{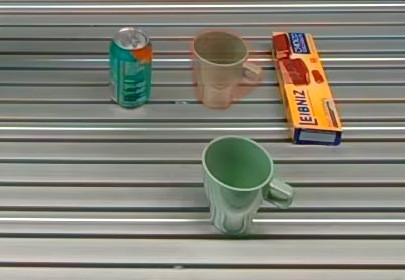}};
    \begin{scope}[shift={(3.63,2.90)}, x=1.62cm, y=1.12cm]\path[tref] (0.270,0.671) -- (0.269,0.667) -- (0.269,0.664) -- (0.269,0.663) -- (0.269,0.661) -- (0.269,0.660) -- (0.269,0.659) -- (0.269,0.658) -- (0.269,0.655) -- (0.269,0.654) -- (0.270,0.652) -- (0.271,0.647) -- (0.272,0.644) -- (0.272,0.644) -- (0.272,0.642) -- (0.273,0.641) -- (0.274,0.639) -- (0.275,0.637) -- (0.275,0.636) -- (0.277,0.634) -- (0.278,0.632) -- (0.280,0.629) -- (0.282,0.627) -- (0.283,0.625) -- (0.286,0.623) -- (0.288,0.620) -- (0.292,0.618) -- (0.294,0.616) -- (0.298,0.614) -- (0.302,0.612) -- (0.304,0.611) -- (0.306,0.610) -- (0.308,0.609) -- (0.315,0.608) -- (0.318,0.608) -- (0.320,0.608) -- (0.326,0.608) -- (0.327,0.608) -- (0.331,0.608) -- (0.332,0.608) -- (0.338,0.610) -- (0.340,0.611) -- (0.343,0.612) -- (0.345,0.613) -- (0.348,0.615) -- (0.352,0.618) -- (0.354,0.619) -- (0.357,0.622) -- (0.359,0.624) -- (0.361,0.626) -- (0.362,0.628) -- (0.363,0.630) -- (0.364,0.631) -- (0.365,0.634) -- (0.366,0.635) -- (0.366,0.636) -- (0.367,0.637) -- (0.368,0.640) -- (0.369,0.642) -- (0.370,0.646) -- (0.371,0.649) -- (0.372,0.654) -- (0.372,0.658) -- (0.372,0.662) -- (0.372,0.663) -- (0.372,0.664) -- (0.372,0.667) -- (0.371,0.668) -- (0.371,0.672) -- (0.369,0.678) -- (0.369,0.678) -- (0.369,0.679) -- (0.367,0.684) -- (0.365,0.687) -- (0.363,0.691) -- (0.361,0.693) -- (0.360,0.695) -- (0.359,0.696) -- (0.358,0.697) -- (0.357,0.698) -- (0.356,0.699) -- (0.354,0.701) -- (0.353,0.702) -- (0.349,0.704) -- (0.344,0.708) -- (0.341,0.709) -- (0.337,0.711) -- (0.332,0.713) -- (0.331,0.713) -- (0.328,0.714) -- (0.324,0.715) -- (0.322,0.715) -- (0.320,0.715) -- (0.318,0.715) -- (0.315,0.715) -- (0.315,0.715) -- (0.312,0.714) -- (0.308,0.714) -- (0.301,0.712) -- (0.300,0.711) -- (0.298,0.711) -- (0.294,0.708) -- (0.292,0.707) -- (0.288,0.704) -- (0.284,0.701) -- (0.281,0.697) -- (0.281,0.697) -- (0.280,0.696) -- (0.279,0.694) -- (0.277,0.692) -- (0.277,0.691) -- (0.277,0.691) -- (0.275,0.688) -- (0.274,0.686) -- (0.273,0.684) -- (0.273,0.683) -- (0.272,0.682) -- (0.272,0.679) -- (0.270,0.675) -- cycle;
\path[tref] (0.669,0.789) -- (0.716,0.509) -- (0.723,0.501) -- (0.724,0.499) -- (0.725,0.499) -- (0.726,0.499) -- (0.734,0.498) -- (0.799,0.495) -- (0.823,0.494) -- (0.826,0.494) -- (0.827,0.494) -- (0.830,0.495) -- (0.831,0.495) -- (0.835,0.497) -- (0.836,0.498) -- (0.835,0.502) -- (0.835,0.503) -- (0.831,0.523) -- (0.826,0.545) -- (0.819,0.578) -- (0.809,0.621) -- (0.789,0.707) -- (0.762,0.824) -- (0.758,0.839) -- (0.758,0.840) -- (0.758,0.840) -- (0.756,0.841) -- (0.747,0.844) -- (0.746,0.844) -- (0.743,0.844) -- (0.727,0.845) -- (0.695,0.847) -- (0.676,0.848) -- (0.667,0.849) -- (0.665,0.849) -- (0.661,0.846) -- (0.661,0.845) -- (0.661,0.843) -- (0.663,0.831) -- (0.664,0.824) -- (0.666,0.810) -- (0.666,0.808) -- cycle;
\draw[tseg] (0.320,0.652) -- (0.739,0.682);
\draw[taxxg] (0.555,0.746) -- (0.415,0.835);
\draw[taxyg] (0.555,0.746) -- (0.579,0.931);
\draw[taxzg] (0.555,0.746) -- (0.614,0.861);
\filldraw[draw=black!65, fill=white, line width=0.25pt] (0.555,0.746) circle[radius=0.45pt];\end{scope}
  \end{scope}
  \draw[draw=black!30, line width=0.4pt, rounded corners=2pt] (3.63,2.90) rectangle ++(1.62,1.12);
\photolabel{text=cgoal!85!black, anchor=north east}{4.44,3.62}{$\hat\Tgoal$}
\node[sym, text=cgoal!85!black, anchor=south] at (4.44,4.06) {Goal Pose};
\draw[garr] (gr.east) -- (3.63,3.46);
\node[blk, font=\scriptsize] (ape) at (2.70,1.80) {Absolute Pose\\Estimator};
\draw[arr, rounded corners=2pt] (0.96,2.90) -- (0.96,1.80) -- (ape.west);
\draw[arr] ([xshift=-0.22cm]gr.south) -- node[left, sym, inner sep=2pt, align=right] {subject\\[-1pt]mask} ([xshift=-0.22cm]gr.south |- ape.north);
\draw[arr] ([xshift=0.22cm]gr.south |- ape.north) -- node[right, sym, inner sep=2pt, text=cpose!80!black] {$\hat\Tobj_0$} ([xshift=0.22cm]gr.south);

\begin{scope}[shift={(5.85,2.85)}]
  \draw[fill=white, draw=black!25, line width=0.5pt, rounded corners=2pt] (0,0) rectangle (2.10,1.90);
  \triad{0.48}{0.45}{15}{s}{0.42}
  \triad{1.58}{1.28}{55}{g}{0.42}
  \draw[parr, shorten <=3pt, shorten >=3pt] (0.48,0.45) -- (1.58,1.28);
  \node[sym, text=cpose!75!black, anchor=north west] at (0.98,0.86) {$\gtok_t$};
  \node[sym, text=cpose!80!black, anchor=north west] at (0.52,0.40) {$\hat\Tobj_t$};
  \node[sym, text=cgoal!85!black, anchor=north west] at (1.63,1.25) {$\hat\Tgoal$};
\end{scope}
\draw[garr] (5.25,3.46) -- (5.85,3.46);
\node[blk, fill=cpose!14, draw=cpose!80!black, text width=1.20cm, minimum height=1.2cm, inner xsep=2pt] (tok) at (8.98,3.80) {Pose\\Tokenizer};
\draw[parr] (7.95,3.80) -- (tok.west);
\node[gsym, anchor=north west, align=left] at (7.25,2.78) {$\qtok_t$: current pose\\[-1pt] $\gtok_t$: goal residual};

\node[sym, anchor=west] (instr2) at (10.25,0 |- instruction.east) {instruction $\instr$};
\node[sym, anchor=base] (external) at (12.77,0 |- instr2.base) {external $\Iext_t$};
\node[sym, anchor=base] (wrist) at (14.46,0 |- instr2.base) {wrist $\Iwr_t$};
\node[sym, anchor=base] (state) at (16.20,0 |- instr2.base) {robot state $\rstate_t$};
\node[blk, fill=black!8, draw=black!50, minimum width=7.05cm, minimum height=0.34cm, inner sep=1.5pt] (prefix) at (13.575,4.38) {VLM context tokens};
\draw[arr] (external.south) -- (external.south |- prefix.north);
\draw[arr] (wrist.south) -- (wrist.south |- prefix.north);
\draw[arr] (instr2.south) -- (instr2.south |- prefix.north);
\draw[arr] (state.south) -- (state.south |- prefix.north);
\draw[arr] (4.76,0 |- instruction.east) -- (instr2.west);
\node[tk, fill=cpose!55] (cq2) at (10.40,3.80) {};
\node[tk, fill=cpose!55] (cg2) at (10.66,3.80) {};
\draw[black!45, line width=0.4pt] (10.85,3.62) -- (10.85,3.98);
\foreach \i in {0,...,4}{\node[tk, fill=cact!25] (a\i) at (11.06+\i*0.26,3.80) {};}
\node[sym] at (12.40,3.80) {$\cdots$};
\node[tk, fill=cact!25] (a9) at (12.68,3.80) {};
\node[sym, text=cpose!80!black, anchor=south] at (10.53,3.96) {$\cq\,\cg$};
\node[sym, text=cact!55!black, anchor=west, inner sep=0pt] at (12.88,3.80) {$\mathbf{x}^A_\tau$};
\draw[fill=white, draw=black!30, line width=0.5pt, rounded corners=1.5pt] (10.42,2.58) rectangle (13.26,3.44);
\draw[fill=white, draw=black!30, line width=0.5pt, rounded corners=1.5pt] (10.34,2.50) rectangle (13.18,3.36);
\draw[fill=white, draw=black!65, line width=0.5pt, rounded corners=1.5pt] (10.26,2.42) rectangle (13.10,3.28);
\node[sym, align=center] at (11.68,2.85) {Action Expert\\[-1pt]{\color{black!55}adaRMSNorm}\\[-1pt]$\temb{+}\modu_t$};
\node[gsym, anchor=west, inner sep=0pt] at (13.30,2.52) {$\times L$};
\draw[arr] (12.10,4.21) -- (12.10,3.97);
\draw[arr] (11.70,3.62) -- (11.70,3.44);
\draw[parr] (tok.east) -- (9.95,3.80) -- (cq2.west);
\draw[parr] (tok.south) |- node[pos=0.75, above, sym, text=cpose!80!black] {$\modu_t$} (10.26,2.85);
\draw[arr] (13.10,2.85) -- (13.60,2.85);
\begin{scope}[shift={(13.63,2.70)}]
  \foreach \i in {0,...,9}{\pgfmathsetmacro{\shade}{\i<5 ? 60 : 22}\fill[cact!\shade, draw=cact!70!black, line width=0.3pt] (\i*0.105,0) rectangle ++(0.085,0.30);}
\end{scope}
\node[sym, anchor=south] at (14.15,3.02) {$\Achunk_t$};
\node[blk, fill=black!7, draw=cact!70!black] (rob) at (16.35,2.85) {Robot\\Action};
\draw[arr] (14.70,2.85) -- node[above, gsym] {first $K$} (rob.west);
\node[blk, fill=cpose!12, draw=cpose!80!black] (trk) at (13.65,1.80) {Relative Pose\\Tracker};
\draw[arr, rounded corners=2pt] (rob.south) -- (16.35,1.80) -- (trk.east);
\node[gsym, anchor=north, align=center, inner sep=1.5pt] at (15.62,1.69) {new RGB-D,\\[-1pt]gripper kinematics};
\draw[parr, rounded corners=2pt] (trk.west) -- (6.90,1.80) -- (6.90,2.85);
\node[sym, text=cpose!80!black, anchor=east, inner sep=2pt] at (6.84,2.32) {$\hat\Tobj_{t+K}$};
\end{tikzpicture}
\caption{\textbf{\method.}
(a)~At command time, the goal reasoner grounds the instruction $\instr$ in the external RGB-D image, the pose estimator~\cite{Li_2026_CVPR} returns the start pose $\hat\Tobj_0$ from the subject mask, and the reasoner composes the goal pose $\hat\Tgoal$ (the goal image is a composite for illustration).
(b)~The pose tokenizer maps the current pose $\qtok_t$ and the residual $\gtok_t$ to two tokens $\cq,\cg$ and a modulation vector $\modu_t$.
(c)~The action expert attends to the context tokens of the vision-language model and to $\cq,\cg$, and $\modu_t$ shifts the flow-time embedding $\temb$ that conditions the normalization of every layer. After the first $K$ actions, the relative pose tracker updates the pose, and the loop repeats.}
\vspace{-8pt}
\label{fig:method}
\end{figure*}

As shown in Fig.~\ref{fig:teaser}, we propose \method, using object pose as a shared representation for language goals, action conditioning, and execution feedback.
Built on \pizerofive~\cite{intelligence2025pi05}, \method conditions its action expert on two continuous pose tokens (\cref{fig:method}): the commanded object's current pose and the translation and rotation remaining to the goal.
Both tokens are updated from the tracked pose after each action chunk, allowing subsequent actions to account for the motion actually achieved.

This feedback begins with a semantic goal reasoner, which grounds the subject and reference objects in a calibrated RGB-D observation.
It combines learned relational placement with geometric primitives for stated distances and angles to form an $\SE$ goal pose.
During execution, a CAD-free relative pose tracker fuses visual point tracks from TAPNext++~\cite{tapnextpp} with gripper kinematics, using kinematics to maintain the estimate through occlusion and visual measurements to correct deviations such as in-hand slip.
The goal remains fixed, while the tracked pose continually updates the feedback used by the policy.

Related approaches connect geometry and policy learning through pose pretraining~\cite{lin2026posevla}, metric end-effector--object interaction modeling~\cite{wang2026actinginmeters}, or refreshed 3D object records for action conditioning and verification~\cite{ren2026potvla}.
Within this line of work, \method focuses on the translation and rotation remaining to a language-specified object goal, using the updated residual to guide metric instructions, semantic spatial relations, and their compositions.

To evaluate this formulation, we introduce a simulation and a real-world benchmark, both scored by the achieved object pose.
\simbench extends LIBERO~\cite{liu2023libero} with \prov{15k} demonstrations and 540 evaluation tasks covering metric displacements down to 1.5\,cm, rotations down to 8$^\circ$, spatial relations, and compositions.
\realbench (\realbenchfull) provides \prov{2,288} demonstrations and \prov{121} test cases across \prov{24} household objects on a Franka Research~3 arm, with scanned digital twins supporting object-pose evaluation.
On \simbench, \method achieves \prov{71.3}\% overall success, compared with \prov{4.4}\% for \pizerofive and \prov{17.4}\% for the strongest of seven learned baselines.
On the real robot, it satisfies \prov{88} of \prov{121} commands, compared with \prov{29} for \pizerofive.

Our contributions are:
\begin{itemize}
\item \textbf{Metric-level object manipulation for VLAs.} \method connects language-specified object goals with dynamically updated pose feedback, enabling one policy to follow semantic, metric, and composed commands.
\item \textbf{A closed-loop pose-aware framework.} A goal reasoner grounds instructions into object goal poses, pose tokens condition actions on the current pose and goal residual, and an integrated CAD-free tracker updates this feedback during execution.
\item \textbf{Pose-based task evaluation in simulation and the real world.}
We introduce a simulation benchmark with \prov{15k} demonstrations and 540 test tasks, and a real-robot benchmark with \prov{2,288} demonstrations and \prov{121} test cases across \prov{24} household objects; both evaluate commands using the annotated object poses.
\end{itemize}

\section{Related Work}
\label{sec:related}
\paragraph{Scene-Centric VLAs and World Action Models.}
Vision-language-action (VLA) models generate actions from pretrained vision-language representations using discrete tokens~\cite{zitkovich2023rt2,kim2024openvla} or continuous action chunks~\cite{chi2023diffusion,octo2024,black2024pi0,intelligence2025pi05,kim2025openvlaoft}, while world action models additionally predict future observations~\cite{guo2024pad,yuan2026fastwam,kim2026cosmospolicy}.
While Internet-scale pretraining endows these models with broad semantic coverage, their control loop remains \emph{scene-centric}.
Whether operating on 2D camera streams or whole-scene 3D point clouds~\cite{zhen2024threedvla,qu2025spatialvla,sun2025geovla,chen2026pointact}, they condition action generation on a global encoding of the entire environment, in which the state of the commanded object remains implicit.
Even when augmented with visual traces~\cite{zheng2025tracevla}, affordance points~\cite{yuan2024robopoint}, or latent action spaces~\cite{lin2026lit}, scene-centric policies lack an explicit metric representation of the commanded object.
\method instead conditions action generation on the 3D object pose and its residual to the goal, enabling VLA policies to follow semantic, metric, and composed commands.

\paragraph{Object-Centric Manipulation and Planning.}
Object-centric methods place a representation of the manipulated object between perception and control.
One line of work guides policies with point tracks, optical flow, or predicted object trajectories~\cite{wen2024atm,bharadhwaj2024track2act,xu2024im2flow2act,hsu2025spot}, and another conditions the policy backbone on object-centric visual prompts~\cite{zhu2025objectvla,li2025controlvla}.
A third addresses language-conditioned spatial rearrangement and derives target placements~\cite{liu2023structdiffusion,zhao2025anyplace}, start-to-goal transformations~\cite{heng2025imagine2act}, visual goal predictions~\cite{chen2025goalvla}, or trajectory optimization constraints~\cite{huang2023voxposer,huang2024rekep,berenson2011tsr}. These approaches expose object-centric motion or target structure; our focus is the changing metric error between the tracked object pose and its language-specified goal.
Most related to our goal reasoning, Open6DOR~\cite{ding2024open6dor} and SoFar~\cite{qi2025sofar} translate semantic spatial instructions into 3D positions and orientations.
However, these systems run open-loop: they chain grasp candidate generation (\eg, GSNet~\cite{wang2021graspness}), inverse kinematics, and motion planning (\eg, OMPL~\cite{sucan2012ompl}) into a precomputed trajectory and execute it without re-observing the object.
Their precomputed execution does not explicitly update the object-to-goal residual from tracked object poses. Metric-VLA supplies this residual to the action expert after each action chunk.
\method closes this loop with a \emph{relative pose tracker} that fuses visual point tracks with gripper kinematics to re-estimate the 6D object pose through grasping and occlusion, so the policy can compensate for slip and disturbances during execution.

\paragraph{Object Pose in Learned Policies.}
Object pose provides a compact $\SE$ state in which metric goal errors can be computed explicitly.
While 3D pose estimation has matured on visual benchmarks~\cite{collet2011moped,xiang2018posecnn,tremblay2018dope,wang2019densefusion,jung2024housecat6d,zhang2024omni6dpose,wen2024foundationpose,Li_2026_CVPR}, learned policies rarely use it as a feedback signal during execution.
Pose-VLA~\cite{lin2026posevla} pretrains on 3D pose as an auxiliary task but drops this supervision during action fine-tuning, after which pose is represented only implicitly in the features.
Other policies predict intermediate poses open-loop without tracking~\cite{su2024mba}, model object-relative end-effector trajectories jointly with actions~\cite{wang2026actinginmeters}, or refresh metric object records for action conditioning and geometric verification~\cite{ren2026potvla}. Metric-VLA integrates language-grounded $\SE$ goal reasoning with visual-kinematic object tracking. After every action chunk, the tracked pose updates two continuous \emph{object pose tokens} encoding the object's current pose and its goal residual, directly grounding subsequent actions in the motion still required.
\section{Metric-VLA}
\label{sec:method}

\subsection{Problem Formulation}
\label{sec:setup}
\begin{figure}[t]
\centering
\begin{tikzpicture}[
  font=\scriptsize, x=1cm, y=1cm,
  hdr/.style={font=\footnotesize\bfseries, anchor=base west, inner sep=0pt},
  gsym/.style={font=\scriptsize, inner sep=1pt, text=black!55},
  lab/.style={font=\scriptsize, inner sep=1pt},
  word/.style={font=\scriptsize\itshape, inner sep=1pt},
  ax/.style={-{Stealth[length=1.5mm,width=1.2mm]}, line width=0.8pt, line cap=round},
  axx/.style={ax, draw=red!75!black},
  axy/.style={ax, draw=green!55!black},
  opt/.style={draw=black!45, dash pattern=on 1.6pt off 1.2pt, line width=0.4pt},
  obj/.style={draw=black!45, fill=black!10, line width=0.4pt},
  start/.style={draw=cpose!75!black, fill=cpose!45, line width=0.5pt},
  goal/.style={draw=cgoal!90!black, fill=cgoal!14, line width=0.6pt, dash pattern=on 1.2pt off 0.8pt},
  mv/.style={-{Stealth[length=1.2mm,width=1.0mm]}, line width=0.7pt, draw=black!85},
]
\newcommand{\zdot}[1]{\draw[draw=blue!75!black, fill=white, line width=0.6pt] (#1) circle (0.075); \fill[blue!75!black] (#1) circle (0.024);}
\newcommand{\triadxy}[3]{\draw[axx] (#1) -- ++(#2:#3); \draw[axy] (#1) -- ++(#2+90:#3); \zdot{#1}}
\newcommand{\camglyph}[1]{\begin{scope}[shift={(#1)}]
  \fill[black!70, rounded corners=0.6pt] (-0.17,-0.25) rectangle (0.17,-0.03);
  \fill[black!70] (-0.07,-0.04) -- (-0.13,0.09) -- (0.13,0.09) -- (0.07,-0.04) -- cycle;
\end{scope}}
\newcommand{\robotglyph}[1]{\fill[black!22] (#1) circle (0.20); \draw[black!55, line width=0.5pt] (#1) circle (0.20);}
\newcommand{\panelbg}{\fill[cpose!6, rounded corners=3pt] (-1.75,-0.50) rectangle (2.20,4.28);}
\newcommand{\optaxis}[1]{%
  \draw[opt] (0,0.10) -- (0,0.30);
  \draw[opt] (0,0.44) -- (#1);
  \draw[black!45, line width=0.4pt] (-0.09,0.27) -- (0.09,0.33) (-0.09,0.41) -- (0.09,0.47);}
\newcommand{\cmdobj}[2]{%
  \draw[goal] (#2) circle (0.10);
  \draw[start] (#1) circle (0.10);
  \draw[mv] ($(#1)+(0.13,0)$) -- ($(#2)+(-0.12,0)$);
  \node[lab, anchor=north] at ($(#1)+(0,-0.12)$) {$\Tobj_0$};
  \node[lab, anchor=north] at ($(#2)+(0,-0.12)$) {$\Tgoal$};}

\node[font=\scriptsize, anchor=base, inner sep=0pt] at (2.35,4.42) {$\instr$: \emph{``move the can 20\,cm to the right''}};

\begin{scope}
  \panelbg
  \node[hdr] at (-1.60,3.94) {(a) \simbench};
  \optaxis{0,1.49}
  \camglyph{0,0}
  \node[lab, anchor=west, align=left] at (0.26,-0.10) {camera\\[-1pt]{\color{black!55}(observer)}};
  \draw[obj, rounded corners=0.5pt, rotate around={20:(1.05,0.62)}] (0.90,0.50) rectangle (1.20,0.74);
  \draw[obj] (-0.80,1.95) circle (0.10);
  \cmdobj{-1.00,0.88}{-0.40,0.88}
  \robotglyph{0,2.81}
  \triadxy{0,2.81}{-90}{0.50}
  \node[lab, anchor=east] at (-0.10,2.45) {$x_{\fB}$};
  \node[lab, anchor=west] at (0.56,2.81) {$y_{\fB}$};
  \node[gsym, anchor=south] at (0,3.05) {robot base $\fB$};
  \triadxy{0,1.49}{0}{0.60}
  \node[lab, anchor=north east] at (-0.08,1.43) {$\fO$};
  \node[lab, anchor=north] at (0.36,1.41) {$x_{\fO}$};
  \node[lab, anchor=east] at (-0.10,1.91) {$y_{\fO}$};
  \node[word, anchor=west] at (0.66,1.49) {to the right};
  \node[word, anchor=west] at (0.10,1.95) {farther from you};
\end{scope}

\begin{scope}[shift={(4.25,0)}]%
  \panelbg
  \node[hdr] at (-1.60,3.94) {(b) \realbench};
  \optaxis{-0.05,3.30}
  \camglyph{0,0}
  \node[lab, anchor=west, align=left] at (0.26,-0.10) {camera\\[-1pt]{\color{black!55}(observer)}};
  \begin{scope}[shift={(-1.07,1.26)}, rotate=-1.52]
    \draw[obj, rounded corners=0.5pt, rotate around={-25:(1.95,-0.38)}] (1.80,-0.50) rectangle (2.10,-0.26);
    \draw[obj] (1.40,-0.55) circle (0.10);
    \robotglyph{0,0}
    \triadxy{0,0}{0}{0.50}
  \end{scope}
  \cmdobj{0.20,1.68}{0.80,1.68}
  \node[lab, anchor=west] at (-0.51,1.25) {$x_{\fB}$};
  \node[lab, anchor=east] at (-1.17,1.71) {$y_{\fB}$};
  \node[gsym, anchor=north, align=center] at (-1.20,0.99) {robot\\[-1pt]base $\fB$};
  \triadxy{-0.05,3.30}{0}{0.60}
  \node[lab, anchor=north east] at (-0.13,3.24) {$\fO$};
  \node[lab, anchor=north] at (0.31,3.22) {$x_{\fO}$};
  \node[lab, anchor=east] at (-0.15,3.72) {$y_{\fO}$};
  \node[word, anchor=west] at (0.61,3.30) {to the right};
  \node[word, anchor=west] at (0.05,3.76) {farther from you};
\end{scope}
\end{tikzpicture}
\vspace{-4pt}
\caption{\textbf{Robot base frame $\fB$ and observer frame $\fO$}, seen from above.
Object poses are expressed in $\fB$; directions in language refer to $\fO$ and are rotated into $\fB$ by $\RBO$.
The same instruction $\instr$ thus moves the object along $+y_{\fB}$ in (a) and along $+x_{\fB}$ in (b).}
\label{fig:frames}
\vspace{-12pt}
\end{figure}
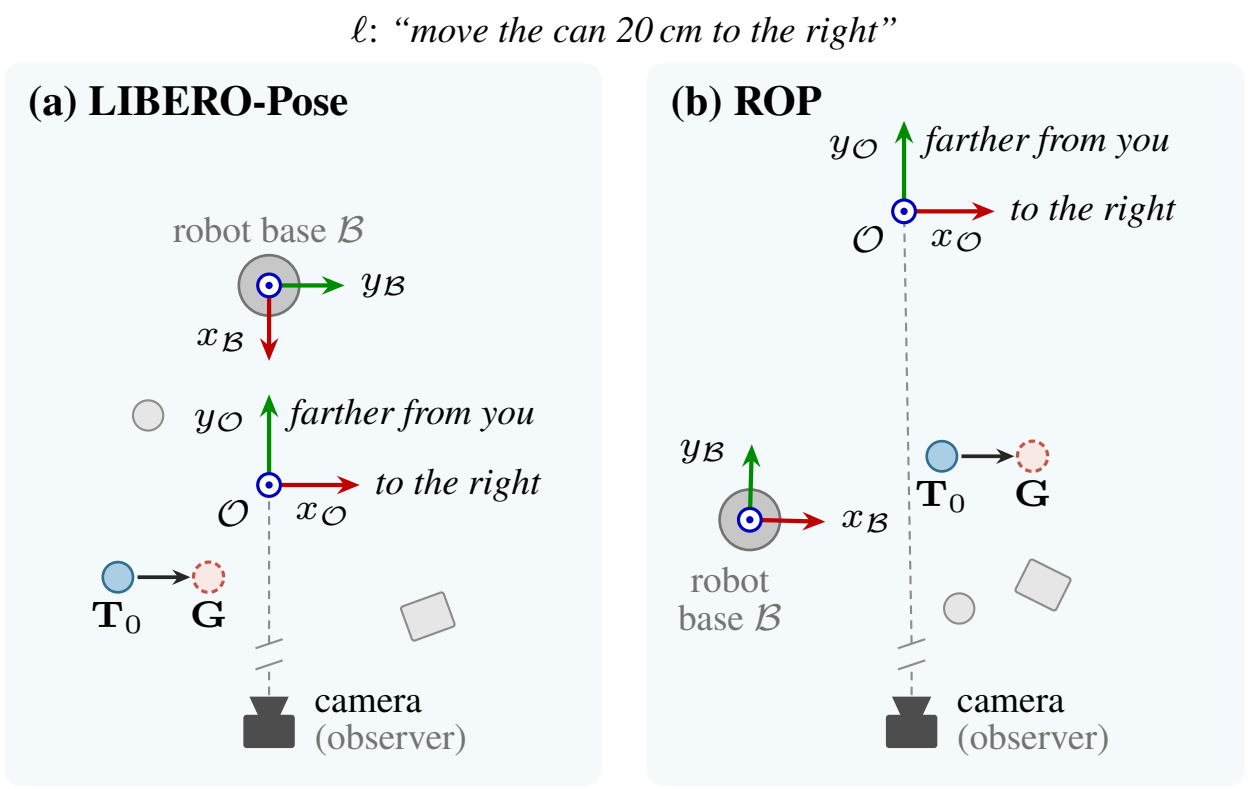

We consider metric object manipulation: a policy receives a language instruction $\instr$ and acts in a scene containing the commanded object. The instruction imposes a geometric condition on the 6D pose $\Tobj_t = (\Rm_t, \pv_t) \in \SE$ in the robot base frame.
A \emph{metric} instruction states a translation or rotation, as in \emph{``move the bottle 10\,cm to the right''} or \emph{``rotate the remote 75 degrees clockwise''}. A \emph{semantic} instruction specifies a relation to other objects or the orientation of a functional part, while a \emph{composed} instruction attaches a metric magnitude to a spatial relation (\cref{sec:benchmark}).
Directions in language refer to an observer frame $\fO$ fixed to the table and aligned with the external camera (\cref{fig:frames}).
The stated centimeters specify a displacement of the object's position from its pose $\Tobj_0$ at command time, and the stated degrees an orientation variation from that initial state.
A relational command requires the object to end in a region defined by reference objects.
An episode succeeds when the final object pose $\Tobj_T$ satisfies these geometric constraints (\cref{sec:metrics}).

At command time, a semantic goal reasoner $f_\phi$ derives the goal pose $\hat\Tgoal$ from the instruction $\instr$, the initial external RGB-D observation $(\Iext_0, \Dext_0)$, and the estimated start pose $\hat\Tobj_0$.
At control step $t$, the policy $\pi$ conditions each action chunk on this goal and the object's current pose $\hat\Tobj_t$:
\begin{equation}
\begin{aligned}
\hat\Tgoal &= f_\phi(\instr, \Iext_0, \Dext_0, \hat\Tobj_0), \\
\Achunk_t &\sim \pi(\cdot \mid \obs_t, \instr, \hat\Tobj_t, \hat\Tgoal).
\end{aligned}
\label{eq:formulation}
\end{equation}
Here, $\obs_t = (\Iext_t, \Iwr_t, \rstate_t)$ comprises the external and wrist RGB images and the robot state; $\Achunk_t$ is an $H$-step action chunk, of which the first $K \le H$ actions execute before the tracker re-estimates $\hat\Tobj_{t+K}$ and the policy replans.
First, a \emph{semantic goal reasoner} grounds the instruction in the scene and returns one object goal pose $\hat\Tgoal$ (\cref{sec:goal}).
Second, for each action chunk, \emph{object pose tokens} provide the action expert with the object's current pose $\hat\Tobj_t$ and the residual still to be executed (\cref{sec:tokens}).
Third, after each action chunk executes, a \emph{relative pose tracker} re-observes the object to update its pose and residual (\cref{sec:loop}).

\subsection{Semantic Goal Reasoning}
\label{sec:goal}

At command time, the goal reasoner $f_\phi$ translates the instruction $\instr$ and initial RGB-D observation $(\Iext_0, \Dext_0)$ into an object goal pose $\hat\Tgoal = (\hat\Rm_G, \hat\pv_G)$ (\cref{fig:method}a).
Patch features from frozen DINOv3~\cite{simeoni2025dinov3} and SigLIP~\cite{zhai2023siglip}, coupled with unprojected 3D coordinates, attend to the T5 instruction embedding~\cite{raffel2020t5}.
From this representation, the reasoner grounds the commanded subject via a patch-level \emph{subject map} and localizes reference entities by their 3D centroids $\mathbf{c}_\text{ref}$.
Prompted by the subject map, SAM\,2~\cite{ravi2024sam2} yields an object mask, from which OPT-Pose~\cite{Li_2026_CVPR} estimates the start pose $\hat\Tobj_0 = (\hat\Rm_0, \hat\pv_0)$. To avoid numeric drift from direct coordinate regression (\cref{tab:goal}), we construct goals from structured \emph{geometric primitives}.
Positions combine a start-pose or reference anchor with an analytic displacement or learned relational placement.
Rotations compose the initial orientation with optional functional-axis alignments and vertical yaws.
For metric commands and directed moves (\eg, \emph{``rotate 45 degrees clockwise''} or \emph{``move 8.9\,cm towards the can''}), this gives:
\begin{equation}
\hat\pv_G = \hat\pv_0 + \sum_k v_k \mathbf{d}_k, \qquad \hat\Rm_G = \Rm_z(\theta^\star)\hat\Rm_0,
\label{eq}
\end{equation}
where magnitudes $v_k$ and yaw angles $\theta^\star$ are extracted from instruction literals, while the network predicts their directional roles.
Each $\mathbf{d}_k \in \R^3$ is a unit direction in the robot base frame $\fB$, obtained by rotating an observer axis through $\RBO$ or pointing toward the reference centroid.

The reasoner is trained on scene--instruction--goal triples using synthetic tabletop object compositions~\cite {zhang2024omni6dpose,jung2024housecat6d}. The multi-task objective combines:
(i)~\emph{grounding losses} that train patch-level subject segmentation and regress 3D reference centroids $\mathbf{c}_\text{ref}$ via smooth-$\ell_1$ loss;
(ii)~\emph{literal parsing losses} that map parsed numeric literals $v_k$ to their geometric roles (\eg, translation magnitude, yaw angle) and predict their directional axes $\mathbf{d}_k$ and signs;
(iii)~\emph{primitive selection loss} via multi-positive cross-entropy over an admissible set $\mathcal{S}$, which includes every configuration whose composed pose satisfies the task's geometric success criteria;
and (iv)~\emph{geometric auxiliary losses} that penalize orientation tilt and enforce exact metric updates on stated quantities. We refer to the supplementary material for further details.

\subsection{Object Pose Tokens}
\label{sec:tokens}

For each action chunk, the action expert requires two types of spatial information: where the object is, and how much motion remains.
We parameterize rotations using the continuous 6D representation~\cite{zhou2019continuity}, where $\rsix(\Rm) \in \R^6$ stacks the first two columns of $\Rm$.
From the current pose $\hat\Tobj_t = (\hat\Rm_t, \hat\pv_t)$ and goal pose $\hat\Tgoal = (\hat\Rm_G, \hat\pv_G)$, we construct two 9D vectors (\cref{fig:method}b):
\begin{equation}
\begin{aligned}
\qtok_t &= \big[\, \hat\pv^{\fO}_t / \pscale, \ \ \rsix(\hat\Rm^{\fO}_t) - \rsix(\mathbf{I}) \,\big], \\
\gtok_t &= \big[\, (\hat\pv_G - \hat\pv_t) / \pscale, \ \ \rsix(\hat\Rm_G \hat\Rm_t^\top) - \rsix(\mathbf{I}) \,\big],
\end{aligned}
\label{eq:tokens}
\end{equation}
where a length scale $\pscale > 0$ normalizes translation, and subtracting $\rsix(\mathbf{I})$ centers identity rotation at zero.
Here, $\qtok_t$ localizes the object in the observer frame $\fO$.
The \emph{goal residual} $\gtok_t$ captures the remaining displacement and relative rotation in the robot base frame $\fB$, aligned with the action space; it shrinks as the object approaches the goal.
When a command leaves orientation unconstrained, a task mask zeroes the rotation entries of $\gtok_t$ to relax the rotation requirement for the policy.

\label{sec:action}
The action expert integrates the pose tokens of \cref{eq:tokens} into action generation (\cref{fig:method}c).
Two MLPs map $\qtok_t$ and $\gtok_t$ to condition tokens $\cq, \cg$ for cross-attention, while a third projects $\gtok_t$ to a modulation $\modu_t$ that is added to the flow-time embedding, $\temb + \modu_t$, which conditions adaRMSNorm~\cite{intelligence2025pi05} in all transformer blocks.

\subsection{Relative Pose Tracker}
\label{sec:loop}

The goal pose $\hat\Tgoal$ is established at command time, while the current pose $\hat\Tobj_t$ is updated after each action chunk to close the feedback loop (\cref{fig:method}c).
While the policy executes $K$ actions per chunk, relative tracking operates across intermediate camera frames at 10\,Hz.
During manipulation, neither modality suffices alone: 2D point tracks suffer from severe manipulator occlusions and drift during transport, while kinematics cannot detect slip or track ungrasped objects.
We resolve these complementary failure modes with a CAD-free interacting multiple model (IMM) filter~\cite{blom1988imm}.

\paragraph{Observation Streams.}
Tracking is initialized with the start pose $\hat\Tobj_0$ and the subject mask of \cref{sec:goal}: TAPNext++~\cite{zholus2025tapnext,tapnextpp} tracks dense 2D points inside the mask, which SAM\,2~\cite{ravi2024sam2} propagates across frames to prune tracks leaving the object.
Across intermediate camera frames $n$, two streams update in parallel:
(i)~\emph{Visual measurement}: surviving points are lifted via external metric depth $\Dext_n$, and rigid Kabsch alignment~\cite{kabsch1976solution} with RANSAC~\cite{fischler1981ransac} yields a relative pose measurement $\mathbf{Z}_n \in \SE$ with respect to the initial object frame; to replenish points lost to occlusions, new tracks are periodically sampled from the mask during confident fits and registered with the filtered pose;
(ii)~\emph{Kinematic propagation}: robot forward kinematics provides the end-effector pose $\mathbf{E}_n \in \SE$, its motion increment $\mathbf{E}_n\mathbf{E}_{n-1}^{-1}$, and the scalar gripper opening $\gamma_n$.

\paragraph{Adaptive Kinematic-Visual Fusion.}
An IMM filter fuses these streams across three physical motion modes, attached, static, and free ($m \in \{\mathrm{att}, \mathrm{stat}, \mathrm{free}\}$):
\begin{equation}
\Tobj_n = \mathbf{M}^{(m)}_n\,\Tobj_{n-1}\,\mathrm{Exp}(\mathbf{w}), \quad \mathbf{w}\sim\mathcal{N}(\mathbf{0},\mathbf{Q}_m),
\label{eq:modes}
\end{equation}
where $\mathbf{M}^{(\mathrm{att})}_n = \mathbf{E}_n\mathbf{E}_{n-1}^{-1}$ carries the object rigidly with the gripper, while $\mathbf{M}^{(\mathrm{stat})}_n = \mathbf{M}^{(\mathrm{free})}_n = \mathbf{I}$.
Static mode enforces zero motion with minimal process noise $\mathbf{Q}_\mathrm{stat}$ (active before grasp or after release), while free mode uses inflated covariance $\mathbf{Q}_\mathrm{free}$ to accommodate in-hand slip or table sliding.
Transition probabilities $\pi_{im}$ between modes are smoothly modulated by the gripper opening $\gamma_n$, and mode probabilities update as $\mu_n^{(m)} \propto \big(\sum_i \pi_{im}(\gamma_n)\mu_{n-1}^{(i)}\big)\,p(\mathbf{Z}_n\mid m)$~\cite{blom1988imm}, where the measurement likelihood $p$ is a robust Student-$t$ density that downweights noisy fits without hard rejection.
The posterior pose $\hat\Tobj_n$ is the mode-weighted mean.
This adaptive interplay leverages the complementary strengths of both modalities: kinematics anchors motion under heavy manipulator occlusions, visual evidence corrects physical slip, and the static mode prevents phantom motion during missed grasps.
After executing $K$ steps, the latest filtered pose $\hat\Tobj_{t+K}$ updates the state, refreshing the goal residual $\gtok_{t+K}$ in \cref{eq:tokens} to close the control loop.
\section{Simulation and Real-World Benchmarks}
\label{sec:benchmark}

\begin{figure*}[t]
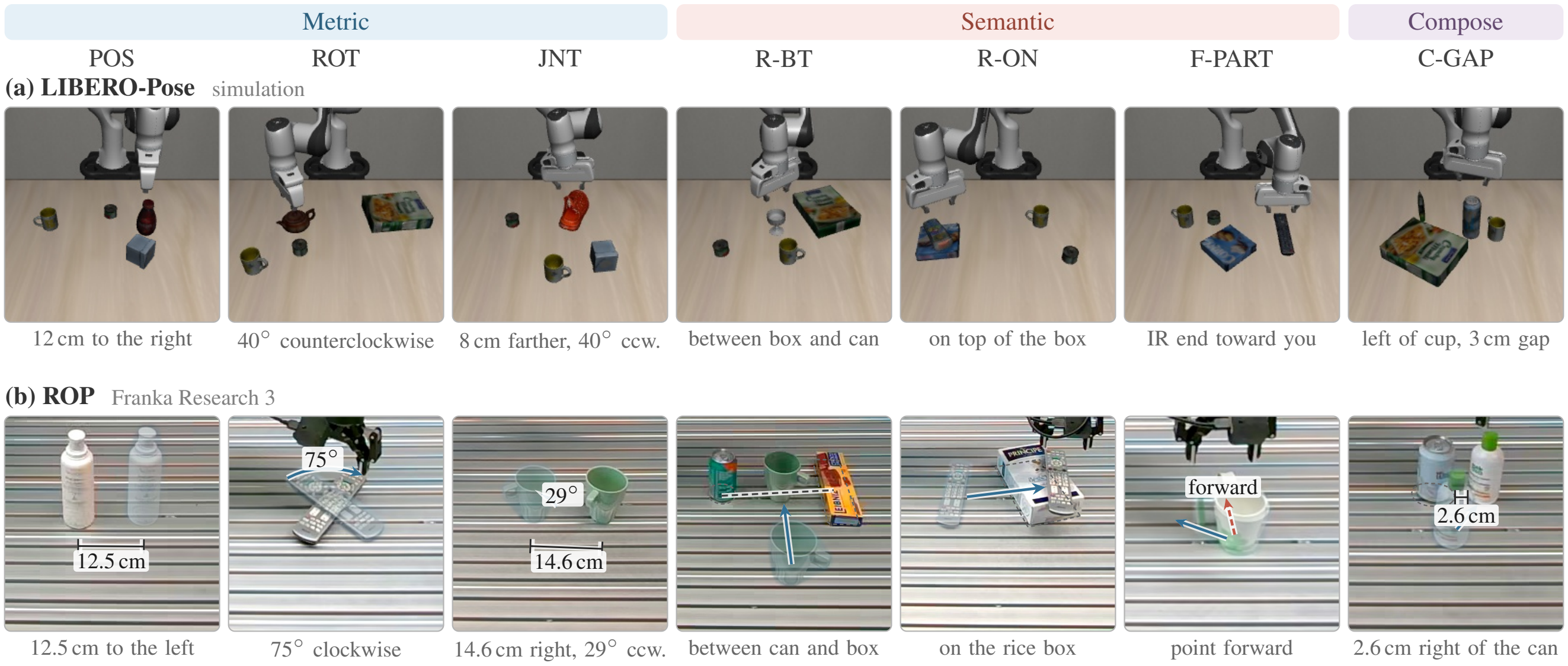

\centering
\providecolor{cmot}{RGB}{153,112,171}%
\begin{tikzpicture}[
  font=\scriptsize, x=1cm, y=1cm,
  hdr/.style={font=\footnotesize\bfseries, anchor=north west, inner sep=0pt},
  fact/.style={font=\scriptsize, anchor=north west, inner sep=0pt, text=black!70, align=left},
  capt/.style={font=\scriptsize, text=black!60, align=center, anchor=north, inner sep=1pt, execute at begin node={\hyphenpenalty=10000\exhyphenpenalty=10000}},
  chip/.style={font=\scriptsize, inner sep=1.3pt, text=white, fill=black!65, fill opacity=0.9, text opacity=1, rounded corners=1pt},
  halo/.style={preaction={draw=white, line width=#1, opacity=0.8}}, halo/.default=2pt,
  tref/.style={draw=black!60, dash pattern=on 1.2pt off 0.9pt, line width=0.45pt},
  tarrow/.style={-{Stealth[length=1.5mm,width=1.3mm]}, line width=0.85pt, draw=cpose!75!black, halo},
  tdim/.style={{Bar[width=1.6mm]}-{Bar[width=1.6mm]}, line width=0.55pt, draw=black!85, halo=1.6pt},
  tarc/.style={-{Stealth[length=1.4mm,width=1.2mm]}, line width=0.85pt, draw=cpose!75!black, halo},
  tseg/.style={dash pattern=on 2pt off 1.2pt, line width=0.6pt, draw=black!75, halo=1.6pt},
  tdirstart/.style={-{Stealth[length=1.3mm,width=1.1mm]}, line width=0.85pt, draw=cpose!75!black, halo},
  tdirgoal/.style={-{Stealth[length=1.3mm,width=1.1mm]}, line width=0.85pt, draw=cgoal!90!black, dash pattern=on 2pt off 1pt, halo},
  tlab/.style={font=\scriptsize, fill=white, fill opacity=0.9, text opacity=1, inner sep=1pt, rounded corners=1pt, text=black!90},
]
\newcommand{\photocard}[5]{%
  \begin{scope}
    \clip[rounded corners=2pt] (#1,#2) rectangle ++(#3,#4);
    \node[inner sep=0pt, anchor=south west] at (#1,#2) {\includegraphics[width=#3cm,height=#4cm]{figures/teaser/#5.jpg}};
    \begin{scope}[shift={(#1,#2)}, x=#3cm, y=#4cm]\input{figures/teaser/#5_overlay}\end{scope}
  \end{scope}
  \draw[draw=black!30, line width=0.4pt, rounded corners=2pt] (#1,#2) rectangle ++(#3,#4);}
\newcommand{\plaincard}[5]{%
  \begin{scope}
    \clip[rounded corners=2pt] (#1,#2) rectangle ++(#3,#4);
    \node[inner sep=0pt, anchor=south west] at (#1,#2) {\includegraphics[width=#3cm,height=#4cm]{figures/teaser/#5.jpg}};
  \end{scope}
  \draw[draw=black!30, line width=0.4pt, rounded corners=2pt] (#1,#2) rectangle ++(#3,#4);}
\def\tw{2.40}\def\gap{0.10}\def\xa{0.00}
\tikzset{colhdr/.style={font=\footnotesize, text=black!75, anchor=south, inner sep=1.5pt},
         grp/.style={font=\footnotesize, anchor=base, inner sep=0pt},
         rowtag/.style={font=\footnotesize\bfseries, anchor=south west, inner sep=0pt, text=black!80},
         rowsub/.style={font=\scriptsize, anchor=south west, inner sep=0pt, text=black!50}}
\tikzset{band/.style={rounded corners=2.5pt, draw=none}}
\fill[band, cpose!14] (\xa,7.00) rectangle (\xa+3*\tw+2*\gap,7.40);
\fill[band, cgoal!13] (\xa+3*\tw+3*\gap,7.00) rectangle (\xa+6*\tw+5*\gap,7.40);
\fill[band, cmot!16] (\xa+6*\tw+6*\gap,7.00) rectangle (\xa+7*\tw+6*\gap,7.40);
\node[grp, text=cpose!65!black] at (\xa+1.5*\tw+1*\gap,7.115) {Metric};
\node[grp, text=cgoal!70!black] at (\xa+4.5*\tw+4*\gap,7.115) {Semantic};
\node[grp, text=cmot!65!black] at (\xa+6.5*\tw+6*\gap,7.115) {Compose};
\foreach \i/\t in {0/POS, 1/ROT, 2/JNT, 3/{R-BT}, 4/{R-ON}, 5/{F-PART}, 6/{C-GAP}}{%
  \node[colhdr] at (\xa+\i*\tw+\i*\gap+\tw/2, 6.64) {\t};}
\def\ya{3.85}
\node[rowtag] (ta) at (\xa,\ya+\tw+0.08) {(a) \simbench};
\node[rowsub, right=5pt of ta.east, anchor=west] {simulation};
\foreach \i/\n in {0/lp_pos_bottle_last, 1/lp_rot_last, 2/lp_jnt_last, 3/lp_rbt_last, 4/lp_ron_last, 5/lp_fpart_last, 6/lp_cgap_last}{%
  \plaincard{\xa+\i*\tw+\i*\gap}{\ya}{\tw}{\tw}{\n}}
\foreach \i/\c in {0/{12\,cm to the right}, 1/{40$^\circ$ counterclockwise}, 2/{8\,cm farther, 40$^\circ$ ccw.}, 3/{between box and can}, 4/{on top of the box}, 5/{IR end toward you}, 6/{left of cup, 3\,cm gap}}{%
  \node[capt, text width=2.38cm] at (\xa+\i*\tw+\i*\gap+\tw/2, \ya-0.05) {\c};}
\def\yb{0.40}
\node[rowtag] (tb) at (\xa,\yb+\tw+0.08) {(b) \realbench};
\node[rowsub, right=5pt of tb.east, anchor=west] {Franka Research 3};
\foreach \i/\n in {0/sq_pos, 1/sq_rot, 2/sq_jnt, 3/sq_rbt, 4/sq_ron, 5/sq_fpart, 6/sq_rlr}{%
  \begin{scope}
    \clip[rounded corners=2pt] (\xa+\i*\tw+\i*\gap,\yb) rectangle ++(\tw,\tw);
    \node[inner sep=0pt, anchor=south west] at (\xa+\i*\tw+\i*\gap,\yb) {\includegraphics[width=\twcm,height=\twcm]{figures/teaser/\n.jpg}};
    \begin{scope}[shift={(\xa+\i*\tw+\i*\gap,\yb)}, x=\twcm, y=\twcm]\input{figures/teaser/\n_overlay}\end{scope}
  \end{scope}
  \draw[draw=black!30, line width=0.4pt, rounded corners=2pt] (\xa+\i*\tw+\i*\gap,\yb) rectangle ++(\tw,\tw);}
\foreach \i/\c in {0/{12.5\,cm to the left}, 1/{75$^\circ$ clockwise}, 2/{\makebox[0pt][c]{14.6\,cm right, 29$^\circ$ ccw.}}, 3/{between can and box}, 4/{on the rice box}, 5/{point forward}, 6/{2.6\,cm right of the can}}{%
  \node[capt, text width=2.38cm] at (\xa+\i*\tw+\i*\gap+\tw/2, \yb-0.05) {\c};}
\end{tikzpicture}
\caption{\textbf{Representative tasks from \simbench and \realbench.} The seven columns show task types shared by both benchmarks, covering metric translations and rotations, semantic spatial relations and functional orientations, and combinations of semantic and metric constraints. (a) Final states from teacher demonstrations in \simbench. (b) Final states from recorded demonstrations on a Franka Research~3 arm in \realbench, with translucent overlays marking initial object poses and annotations illustrating the demonstrated motion or spatial relation. Task descriptions appear below each image (ccw.: counterclockwise; JNT: joint translation and rotation).}
\vspace{-8pt}
\label{fig:benchmarks}
\end{figure*}

Both benchmarks evaluate whether final object poses satisfy commanded displacements and spatial relations, using reference-object geometry where required, under the success criteria of \cref{sec:setup}.
While standard manipulation suites~\cite{liu2023libero,fei2026liberoplus,zhou2025liberopro,wang2026liberox,fang2026liberocf,kim2026liberopara,cui2026liberosafety,li2026liberovpro,zhang2026liberomax} score task completion with logical predicates and Open6DOR~\cite{ding2024open6dor} classifies semantic relations, \simbench and \realbench also command continuous metric targets.
As shown across the columns of \cref{fig:benchmarks}, tasks fall into three broad families:
\emph{metric commands} specify an object's translation or rotation (\eg, the 6\,cm shift and $40^\circ$ turn in row~a);
\emph{semantic commands} establish spatial relations to reference objects or functional orientations (\eg, \emph{``between the box and the can''} or \emph{``IR end toward you''});
and \emph{composed commands} bind metric distances to semantic relations (\eg, the 3\,cm gap and 2.6\,cm offset in the last column).
\simbench covers 12 task types and \realbench contains eight of these; \cref{fig:benchmarks} illustrates seven representative types common to both suites.

\subsection{The Simulation Benchmark: \simbench}

We build \simbench on LIBERO's MuJoCo stack~\cite{liu2023libero}, using a Franka Emika Panda arm at 20\,Hz with $224\times224$ external RGB-D and wrist RGB streams.
Row~(a) of \cref{fig:benchmarks} shows representative final states from teacher rollouts.
To test object control beyond memorized trajectories, we vary training magnitudes continuously over 1--25\,cm in translation and 5--120$^\circ$ in rotation.
We also command multiple goals from the same resting scene, so the policy must use the instruction to determine the outcome.
Directions such as \emph{``to the right''} and \emph{``farther from you''} are defined in the observer frame $\fO$.

For training, we use 50 high-precision scanned household objects~\cite{jung2024housecat6d,zhang2024omni6dpose} from ten categories, together with \prov{155} reference and distractor assets.
A scripted Cartesian impedance teacher generates \prov{15k} demonstrations with ground-truth state.
We evaluate policies on 540 tasks covering 30 objects (ten unseen), six task configurations per object (three metric, two semantic, and one composed), and three unseen initial layouts.
Each rollout runs for 650 control steps.
We score the final resting pose under \cref{sec:metrics}, checking spatial relations through footprint, contact, and axis-alignment predicates on the actual reference poses.

\subsection{The Real-World Benchmark: \realbench}
\label{sec:realbench}

\realbench (\realbenchfull) instantiates eight of these task types on a Franka Research~3 arm with scanned household objects (\cref{fig:benchmarks}b).
The robot observes the scene with a calibrated external RGB-D camera (RealSense D455) and a wrist RGB camera (D405).
A scripted teacher collects demonstrations and resets the scene without human help~\cite{pinto2016supersizing,levine2018grasping}.
It measures the object and its references with a model-based estimator~\cite{wang2026wapr} in the D455 image, picks a reachable grasp defined on the scanned mesh, and carries the object to a sampled goal with an object-pose servo~\cite{hutchinson1996servo} at 10\,Hz, which keeps correcting position and yaw through placement and release.
In this way it records \prov{2,288} demonstrations on \prov{19} manipulated and \prov{12} reference objects; five further objects are held out for evaluation. Offline, we annotate each frame with the pose estimator~\cite{wang2026wapr}, fitted jointly to both views and to a frontal camera (D435if) that the policy does not use.
As the physical setup lacks ground truth, we validate all annotations against four physical consistency checks: in-hand rigidity (\prov{1.0}\,mm and \prov{0.7}$^\circ$ median drift), forward-kinematics agreement (\prov{0.3}\% error, \prov{1.5}\,mm residual), table-plane contact (\prov{1.5}\,mm), and multi-view consensus (\prov{6.9}\,mm and \prov{2.1}$^\circ$), automatically filtering \prov{310} corrupt or slipping trials out of \prov{2,665} recordings.
The benchmark evaluates policies on a standardized suite of \prov{121} test cases (\prov{20} on unseen objects), scored on final resting poses and metric tolerances (2\,cm, $30^\circ$; \cref{sec:metrics}).
Commands below 5\,cm or $30^\circ$ approach physical measurement uncertainty and are evaluated exclusively in simulation.

\section{Experiments}
\label{sec:experiments}

We evaluate \method in simulation on \simbench and on a real robot on \realbench.
We first compare \method with state-of-the-art scene-centric VLAs and a modular planner on metric, semantic, and composed tasks (\cref{sec:main}), then ablate and analyze its components (\cref{sec:ablation}).

\begin{table*}[t]
\centering
\caption{\textbf{Object control on \simbench and \realbench.}
Metrics follow \cref{sec:metrics}. \simbench: 540 tasks on 30 objects; Unseen: 180 tasks on 10 objects absent from training. \realbench: \prov{121} physical test cases (\prov{54} metric, \prov{44} semantic, \prov{23} composed; \prov{20} on unseen objects).
Current/Goal: predicted (pred), ground-truth (GT, gray), or no (--) poses. Inputs: external RGB $\Iext$, wrist RGB $\Iwr$, depth $\Dext$.}
\label{tab:main}
\small
\setlength{\tabcolsep}{2.7pt}
\begin{tabular}{@{}l ccc cc ccccc c c c c@{}}
\toprule
& \multicolumn{3}{c}{Input} & \multicolumn{2}{c}{Poses} & \multicolumn{5}{c}{Metric} & Semantic & Compose & All & Unseen \\
\cmidrule(lr){2-4}\cmidrule(lr){5-6}\cmidrule(lr){7-11}
Method & $\Iext$ & $\Iwr$ & $\Dext$ & Current & Goal & SR\up & $e_t$\down & $e_R$\down & ADD-S\up & MSSD\up & SR\up & SR\up & SR\up & SR\up \\
& & & & & & (\%) & (cm) & ($^\circ$) & (\%) & (\%) & (\%) & (\%) & (\%) & (\%) \\
\midrule
\multicolumn{15}{@{}l}{\textit{\simbench} (Simulation)} \\
\pizerofive~\cite{intelligence2025pi05} & \cmark & \cmark & -- & -- & -- & \pz\prov{1.1} & \prov{6.00} & \prov{24.4} & \prov{40.0} & \pz\prov{5.6} & \prov{10.0} & \pz\prov{3.3} & \pz\prov{4.4} & \pz\prov{0.6} \\
OpenVLA-OFT~\cite{kim2025openvlaoft} & \cmark & \cmark & -- & -- & -- & \pz\prov{0.7} & \prov{6.08} & \prov{24.0} & \prov{39.6} & \pz\prov{5.6} & \pz\prov{2.8} & \pz\prov{0.0} & \pz\prov{1.3} & \pz\prov{0.0} \\
Fast-WAM~\cite{yuan2026fastwam} & \cmark & \cmark & -- & -- & -- & \pz\prov{0.4} & \prov{6.96} & \prov{31.9} & \prov{40.0} & \pz\prov{5.9} & \pz\prov{3.3} & \pz\prov{0.0} & \pz\prov{1.3} & \pz\prov{0.0} \\
Cosmos Policy~\cite{kim2026cosmospolicy} & \cmark & \cmark & -- & -- & -- & \pz\prov{3.0} & \prov{6.00} & \prov{21.2} & \prov{37.0} & \pz\prov{5.9} & \prov{39.4} & \prov{16.7} & \prov{17.4} & \pz\prov{7.2} \\
PointACT~\cite{chen2026pointact} & \cmark & -- & \cmark & -- & -- & \pz\prov{0.0} & \prov{8.00} & \prov{39.9} & \prov{38.5} & \pz\prov{5.6} & \pz\prov{1.1} & \pz\prov{0.0} & \pz\prov{0.4} & \pz\prov{0.6} \\
SpatialVLA~\cite{qu2025spatialvla} & \cmark & -- & \cmark & -- & -- & \pz\prov{0.7} & \prov{6.58} & \prov{31.7} & \prov{34.8} & \pz\prov{5.2} & \prov{10.0} & \pz\prov{1.1} & \pz\prov{3.9} & \pz\prov{2.2} \\
GeoVLA~\cite{sun2025geovla} & \cmark & -- & \cmark & -- & -- & \pz\prov{0.7} & \prov{7.99} & \prov{38.9} & \prov{36.7} & \pz\prov{5.6} & \pz\prov{8.3} & \pz\prov{2.2} & \pz\prov{3.5} & \pz\prov{1.7} \\
\ours \textbf{\method (ours)} & \cmark & \cmark & \cmark & pred & pred & \textbf{\prov{67.8}} & \textbf{\prov{0.72}} & \pz\textbf{\prov{4.1}} & \textbf{\prov{84.8}} & \textbf{\prov{57.8}} & \textbf{\prov{78.3}} & \textbf{\prov{67.8}} & \textbf{\prov{71.3}} & \textbf{\prov{60.6}} \\
\midrule
\rowcolor{gray!12} Planner + exec., GT poses~\cite{qi2025sofar} & \cmark & -- & \cmark & GT & GT & \prov{31.5} & \prov{1.61} & \prov{11.0} & \prov{68.1} & \prov{30.0} & \prov{50.0} & \prov{38.9} & \prov{38.9} & \prov{39.4} \\
\rowcolor{gray!12} \method, GT poses & \cmark & \cmark & -- & GT & GT & \prov{75.2} & \prov{0.66} & \pz\prov{1.0} & \prov{86.7} & \prov{74.8} & \prov{87.8} & \prov{83.3} & \prov{80.7} & \prov{64.4} \\
\midrule
\multicolumn{15}{@{}l}{\textit{\realbench} (Franka FR3 Robot)} \\
\pizerofive~\cite{intelligence2025pi05} & \cmark & \cmark & -- & -- & -- & \prov{11.1} & \prov{5.99} & \prov{79.1} & \prov{25.9} & \pz\prov{1.9} & \prov{45.5} & \prov{13.0} & \prov{24.0} & \prov{10.0} \\
\ours \textbf{\method (ours)} & \cmark & \cmark & \cmark & pred & pred & \textbf{\prov{70.4}} & \textbf{\prov{1.52}} & \pz\textbf{\prov{9.6}} & \textbf{\prov{79.6}} & \textbf{\prov{42.6}} & \textbf{\prov{72.7}} & \textbf{\prov{78.3}} & \textbf{\prov{72.7}} & \textbf{\prov{70.0}} \\
\bottomrule
\end{tabular}
\label{tab:main_table}
\vspace{-8pt}
\end{table*}

\subsection{Experimental Setup}
\label{sec:expsetup}

\paragraph{Protocol.}
The 540 evaluation tasks of \simbench comprise 270 metric, 180 semantic, and 90 composed tasks, with 180 tasks instantiated on the ten held-out novel objects (\cref{sec:benchmark}).
All baselines are evaluated on the same tasks and evaluation seed, with one rollout per task.
\realbench evaluates the same command families on a Franka Research~3 arm.%

\paragraph{Implementation Details.}
Simulation policies are fine-tuned on the \prov{15k} teacher demonstrations of \simbench.
\method builds upon \pizerofive~\cite{intelligence2025pi05}: we fine-tuned from the public \pizerofive LIBERO checkpoint for 30k steps with the \pizerofive flow-matching loss on actions, given ground-truth current and goal poses.
Training uses AdamW with a constant learning rate of $5\times10^{-5}$ after 1k warm-up steps and a batch size of 256 on 4 H100 GPUs.
In simulation, the policy operates at 20\,Hz, predicting action chunks with a horizon $H=10$ and executing the first $K=5$ steps before replanning; rollouts use exponential moving average weights (decay 0.999) with $N=10$ flow-sampling steps. On the real robot, \method and \pizerofive are fine-tuned on the \realbench training split, operating at 10\,Hz with $H=50$ and $K=10$.

\paragraph{Baselines.}
We benchmark \method against four representative RGB policies: \textbf{\pizerofive}~\cite{intelligence2025pi05}, \textbf{OpenVLA-OFT}~\cite{kim2025openvlaoft}, \textbf{Fast-WAM}~\cite{yuan2026fastwam}, and \textbf{Cosmos Policy}~\cite{kim2026cosmospolicy}, each fine-tuned from its public checkpoint.
These policies condition on external and wrist RGB images, text instructions, and the robot's proprioceptive state.
To test whether depth alone bridges the metric gap without explicit object modeling (\cref{sec:related}), we also benchmark three RGB-D VLAs: \textbf{PointACT}~\cite{chen2026pointact}, \textbf{SpatialVLA}~\cite{qu2025spatialvla}, and \textbf{GeoVLA}~\cite{sun2025geovla}, which ingest external metric depth alongside RGB.
Each baseline is fine-tuned for 30k steps.
Finally, we compare against an open-loop modular planning baseline built on the execution framework of SoFar~\cite{qi2025sofar,baik2026textguided}. Given ground-truth start and goal poses and the object mask, GSNet~\cite{wang2021graspness} proposes grasps from the initial RGB-D image, and inverse kinematics with OMPL~\cite{sucan2012ompl} plans a staged sequence of approach, grasp, lift, transport, descent, and release, which a joint-position controller executes without observing the object again (Planner + exec.\ in \cref{tab:main}).

\paragraph{Evaluation Metrics.}
\label{sec:metrics}
Success rate (SR) evaluates whether the final pose $\Tobj_T$ satisfies the instruction's geometric constraints (\cref{sec:setup}), with default tolerances of 2\,cm and $30^\circ$ (tightened to 6--9\,mm and 3--5$^\circ$ for small commands). Prior to execution, \emph{goal validity} (\cref{tab:goal}) measures the percentage of predicted goals $\hat\Tgoal$ meeting these same criteria.
On metric tasks, we report median translation and SO(3) errors ($e_t, e_R$) alongside the percentage of rollouts with ADD-S and MSSD below 10\% of the object diameter.

\begin{table*}[t]
\centering
\caption{\textbf{Task breakdown on \realbench and analysis on \simbench.}
\textbf{(a)}~Successful physical rollouts across task families on \realbench (\prov{121} test cases on Franka FR3).
\textbf{(b)}~Pose inputs with ground-truth poses: updated (\cmark), frozen at start (\frozenmark), or removed (--).
\textbf{(c)}~Ground-truth inputs replaced one at a time; GT/GT: simulator poses at every step; track: tracker started from the ground-truth start pose; pred: estimated start pose and predicted goal ($^\dagger$: simulator mask); Track: median tracking error at the end of the rollout.}
\label{tab:ablations}
\begin{subtable}[t]{0.256\textwidth}
\centering
\caption{Real-Robot Task Breakdown (\realbench)}
\label{tab:real_breakdown}
\scriptsize
\setlength{\tabcolsep}{2.0pt}
\begin{tabular}{@{}l r cc@{}}
\toprule
Task family & Cases & \pizerofive & \method \\
\midrule
Translation & 30 & \pz\prov{1} & \textbf{\prov{22}} \\
Rotation \& joint & 24 & \pz\prov{5} & \textbf{\prov{16}} \\
Spatial relations & 35 & \prov{20} & \textbf{\prov{27}} \\
Part orientation & 9 & \pz\prov{0} & \textbf{\pz\prov{5}} \\
Composed & 23 & \pz\prov{3} & \textbf{\prov{18}} \\
\bottomrule
\end{tabular}
\end{subtable}%
\hfill%
\begin{subtable}[t]{0.258\textwidth}
\centering
\caption{Pose Conditioning: SR (\%)}
\label{tab:ablation}
\scriptsize
\setlength{\tabcolsep}{2.0pt}
\begin{tabular}{@{}l ccc >{\hspace{4pt}}c c@{}}
\toprule
Variant & Pose & Res. & Mod. & All & Uns. \\
\midrule
start res. & -- & \frozenmark & \frozenmark & \prov{11.7} & \pz\prov{5.6} \\
frozen start & \frozenmark & \frozenmark & \frozenmark & \prov{26.7} & \prov{12.2} \\
w/o residual & \cmark & -- & -- & \pz\prov{8.1} & \pz\prov{5.0} \\
w/o modul. & \cmark & \cmark & -- & \prov{79.8} & \prov{57.8} \\
\ours \textbf{\method} & \cmark & \cmark & \cmark & \textbf{\prov{80.7}} & \textbf{\prov{64.4}} \\
\bottomrule
\end{tabular}
\end{subtable}%
\hfill%
\begin{subtable}[t]{0.255\textwidth}
\centering
\caption{Closed-Loop Integration}
\label{tab:cells}
\scriptsize
\setlength{\tabcolsep}{2.0pt}
\begin{tabular}{@{}l cccc@{}}
\toprule
Current / Goal & Track\down & SR\up & $e_t$\down & MSSD\up \\
& (mm) & (\%) & (cm) & (\%) \\
\midrule
GT / GT & 0.0 & \prov{80.7} & \prov{0.66} & \prov{74.8} \\
track / GT & \prov{1.8} & \prov{76.7} & \prov{0.73} & \prov{61.5} \\
pred$^\dagger$ / pred & \prov{6.0} & \prov{72.6} & \prov{0.73} & \prov{58.9} \\
\ours pred / pred & \prov{6.1} & \prov{71.3} & \prov{0.72} & \prov{57.8} \\
\bottomrule
\end{tabular}
\end{subtable}
\vspace{-8pt}
\end{table*}

\subsection{Comparison with State-of-the-Art Methods}
\label{sec:main}
\paragraph{Results on \simbench.}
With predicted poses, \method achieves \prov{71.3}\% overall success on \simbench (\cref{tab:main}), substantially outperforming learned baselines (Cosmos Policy \prov{17.4}\%, \pizerofive\ \prov{4.4}\%, others $\le\prov{3.9}\%$).
This advantage spans metric (\prov{67.8}\%), semantic (\prov{78.3}\%), and composed (\prov{67.8}\%) commands, and generalizes to unseen objects (\prov{60.6}\% vs.\ at most \prov{7.2}\% for baselines).
VLAs and WAMs struggle with metric grounding: across 189 translation tasks, they remain within 5\,mm of the start in \prov{85}--\prov{174} rollouts (median errors \prov{6.0}--\prov{8.0}\,cm), with even the top baseline collapsing from \prov{39.4}\% on semantic tasks to \prov{3.0}\% on metric ones. In most rollouts, the gripper stops open above the object and never closes (\prov{358} of 540 for \pizerofive).
The cause is a visual ambiguity: the pose above the object before the grasp looks like the rest pose that ends every demonstration, and for a metric command no image shows where the object started, so the current frame cannot tell a pending command from a completed one, whereas the residual can.
\method delivers fine-grained physical precision (\prov{0.72}\,cm, \prov{4.1}$^\circ$), with \prov{57.8}\% satisfying the MSSD threshold (vs.\ $\le\prov{5.9}\%$).
These gains require closed-loop 3D pose feedback: RGB-D inputs achieve $\le\prov{3.9}\%$, an open-loop planner with ground-truth poses reaches \prov{38.9}\%, and freezing pose tokens drops \method to \prov{26.7}\% (\cref{sec:token_ablation}).
Ground-truth poses provide an upper-bound reference of \prov{80.7}\% (\prov{64.4}\% unseen; \cref{sec:perception}).

\paragraph{Results on \realbench.}
On the physical robot (\cref{tab:main,tab:real_breakdown}), \method satisfies \prov{88} of \prov{121} cases (\prov{72.7}\%), compared to \prov{29} (\prov{24.0}\%) for \pizerofive, while reducing median translation and rotation errors to \prov{1.52}\,cm and \prov{9.6}$^\circ$ (vs.\ \prov{5.99}\,cm and \prov{79.1}$^\circ$).
This precision yields a substantial lead on strict surface metrics (\prov{42.6}\% vs.\ \prov{1.9}\% MSSD).
The breakdown across task families (\cref{tab:real_breakdown}) shows that this advantage concentrates on metric and directional control: \method satisfies \prov{22} of \prov{30} translation cases (vs.\ \prov{1} for \pizerofive), \prov{16} of \prov{24} rotation and joint cases (vs.\ \prov{5}).
Without object pose feedback, \pizerofive lacks physical stopping criteria and drifts arbitrarily during in-hand transport. The gap narrows on qualitative spatial relations without distances (\prov{27} vs.\ \prov{20} of \prov{35}), where semantic affordance suffices. On \prov{20} physical cases on unseen objects, \method succeeds in \prov{14} cases against \prov{2} for \pizerofive.

\subsection{Ablations and Analysis}
\label{sec:ablation}

\paragraph{Pose Token Conditioning.}
\label{sec:token_ablation}
We first evaluate the action expert with ground-truth poses across task families (\cref{tab:ablation}).
Removing the residual token while still updating the current pose reduces SR from \prov{80.7}\% to \prov{8.1}\%, with metric commands collapsing from \prov{75.2}\% to \prov{1.5}\%: current object pose alone cannot guide execution to the goal. Providing only the start residual without the current pose token achieves only \prov{11.7}\% overall (\prov{4.1}\% metric), while freezing both pose and residual at their start reaches \prov{26.7}\% (metric SR falling to \prov{10.4}\%); continuously updating the pose interface substantially improves execution over static start conditioning. Between the two residual conditioning paths (\cref{sec:action}), cross-attention carries most of the signal: removing layer modulation changes SR from \prov{80.7}\% to \prov{79.8}\%, a difference of \prov{0.9} points that is not statistically significant, although success on unseen objects falls from \prov{64.4}\% to \prov{57.8}\%.

\begin{table}[t]
\centering
\caption{\textbf{Perception modules on \simbench.}
\textbf{(a)}~Goals that satisfy the instruction.
\textbf{(b)}~RMSE over all tracker updates.}
\label{tab:perception}
\begin{subtable}[t]{0.600\columnwidth}
\centering
\caption{Goal Reasoner Validity (\%)}
\label{tab:goal}
\scriptsize
\setlength{\tabcolsep}{1.6pt}
\begin{tabular}{@{}l c cccc@{}}
\toprule
Variant & Start & Met. & Sem. & Comp. & All \\
\midrule
w/o primitives & GT & \pz\pz\prov{2.2} & \prov{75.6} & \prov{37.8} & \prov{32.6} \\
w/o learned place. & GT & \prov{100.0} & \prov{81.7} & \prov{73.3} & \prov{89.4} \\
\ours Reasoner & GT & \prov{100.0} & \prov{90.6} & \prov{90.0} & \prov{95.2} \\
\ours Reasoner & pred & \pz\prov{82.6} & \prov{85.6} & \prov{86.7} & \prov{84.3} \\
\bottomrule
\end{tabular}
\end{subtable}
\hfill
\begin{subtable}[t]{0.350\columnwidth}
\centering
\caption{Tracking RMSE\down}
\label{tab:tracker}
\scriptsize
\setlength{\tabcolsep}{2.0pt}
\begin{tabular}{@{}l cc@{}}
\toprule
Modality & trans.\ (mm) & rot.\ ($^\circ$) \\
\midrule
visual only & \prov{73.0} & \prov{26.0} \\
kin.\ (IMM) & \prov{90.2} & \prov{26.7} \\
kin.\ (rigid) & \prov{35.9} & \prov{23.9} \\
\ours IMM fused & \prov{13.8} & \prov{15.1} \\
\bottomrule
\end{tabular}
\end{subtable}
\vspace{-10pt}
\end{table}

\paragraph{Goal Reasoning and Pose Tracking.}
\label{sec:perception_ablation}
\label{sec:perception}
In goal reasoning (\cref{tab:goal}), without geometric primitives, direct regression cannot enforce metric constraints, collapsing metric from \prov{100}\% to \prov{2.2}\% and composed from \prov{90.0}\% to \prov{37.8}\% (\prov{32.6}\% overall). Conversely, replacing the learned placement with fixed heuristics lowers semantic accuracy from \prov{90.6}\% to \prov{81.7}\% (\prov{89.4}\% overall), confirming the necessity of context-aware spatial grounding.
The full reasoner achieves \prov{95.2}\% overall validity under ground-truth starts, and gracefully handles upstream perception noise under predicted starts, retaining \prov{84.3}\% overall (\prov{82.6}\% metric, \prov{86.7}\% composed).
For relative pose tracking (\cref{tab:tracker}), individual sensing fails under physical interaction: visual tracks alone lose the object under occlusion (\prov{73.0}\,mm RMSE), while kinematics alone drifts before contact without visual correction (\prov{90.2}\,mm) or inherits grasp offsets under rigid attachment (\prov{35.9}\,mm). By bridging visual tracking and kinematics, our IMM filter reduces errors to \prov{13.8}\,mm and \prov{15.1}$^\circ$, improving translation accuracy by \prov{61}\% over rigid kinematics and \prov{81}\% over tracking alone.

\paragraph{Closed-Loop System Integration.}
\label{sec:cells_ablation}
\cref{tab:cells} replaces the ground-truth inputs of \method one at a time.
Running the relative tracker from the ground-truth start pose, instead of reading the simulator pose at every step, lowers SR from \prov{80.7}\% to \prov{76.7}\%; the tracked position ends a median \prov{1.8}\,mm from the true one, and MSSD drops more than $e_t$ (\prov{74.8}\% vs.\ \prov{61.5}\%, against \prov{0.66} vs.\ \prov{0.73}\,cm), pointing to orientation drift.
Estimating the start pose, which also anchors the predicted goal, costs another \prov{4.1} points, and predicting the subject mask instead of taking it from the simulator \prov{1.3} more, down to \prov{71.3}\%.
With every input predicted, metric commands keep a median translation error of \prov{0.72}\,cm, and \prov{57.8}\% satisfy the MSSD threshold.

\section{Conclusions and Limitations}
\label{sec:conclusion}

We address the fundamental limitation of VLA policies on precise metric instructions, which leave object pose and progress implicit.
We present \method, pairing $\SE$ goal reasoning with pose-residual tokens and visual--kinematic tracking to turn manipulation into continuous, closed-loop error reduction.
Simulated and real-world experiments demonstrate that object-pose feedback bridges semantic reasoning and metric accuracy, generalizing to unseen objects.
While \method currently focuses on single-stage tasks with static targets, leaving multi-step sequencing as an open challenge, its residual feedback provides the closed-loop primitive for hierarchical planners to chain sub-goals via error convergence.
By closing the loop between language, 3D perception, and action through explicit object state, \method establishes a foundation for generalist robots that translate high-level semantic intent into precise physical reality.
\section*{LLM Usage Statement}
We used LLM assistants for text, figures, and code.
Claude edited the wording, grammar, and length of author-written drafts in all sections, checked the consistency of terms, notation, and numbers, and helped format the bibliography.
Following the authors' instructions, Claude and Codex also helped write the figure-rendering scripts and the training, evaluation, and data-conversion scripts.
All reported numbers come from the authors' own runs, the authors checked every cited reference, and the authors take full responsibility for the paper.

{\small
\bibliographystyle{ieeenat_fullname}
\bibliography{references}
}

\makeatletter
\immediate\write\@auxout{\string\newlabel{main:lastfigure}{{\the\c@figure}{}{}{}{}}}
\immediate\write\@auxout{\string\newlabel{main:lasttable}{{\the\c@table}{}{}{}{}}}
\immediate\write\@auxout{\string\newlabel{main:lastequation}{{\the\c@equation}{}{}{}{}}}
\makeatother

\end{document}